\documentclass[10pt,twocolumn,letterpaper]{article}

\usepackage[pagenumbers]{iccv} 

\definecolor{iccvblue}{rgb}{0.21,0.49,0.74}
\usepackage[pagebackref,breaklinks,colorlinks,allcolors=iccvblue]{hyperref}

\newcommand*{\affaddr}[1]{#1} 
\newcommand*{\affmark}[1][*]{\textsuperscript{#1}}
\newcommand*{\email}[1]{\texttt{#1}}

\def\paperID{3386} 
\def\confName{ICCV}
\def\confYear{2025}

\title{Temporal Regularization Makes Your Video Generator Stronger}

\author{%
Harold Haodong Chen\affmark[1,2]~~~ 
Haojian Huang\affmark[1,4]~~~
Xianfeng Wu\affmark[1,2]~~~
Yexin Liu\affmark[1,2]~~~\\
Yajing Bai\affmark[1,2]~~~
Wen-Jie Shu\affmark[1,2]~~~
Harry Yang\affmark[1,2]~~~
Ser-Nam Lim\affmark[1,3]~~~\\
\affaddr{\affmark[1]Everlyn AI~~~~~}
\affaddr{\affmark[2]HKUST~~~~~}
\affaddr{\affmark[3]UCF~~~~~}
\affaddr{\affmark[4]HKU~~~~~}\\
\small\textbf{\email{Project page:\textcolor{magenta}{\url{https://haroldchen19.github.io/FluxFlow/}}}}
}

\begin{document}
\twocolumn[{
\renewcommand\twocolumn[1][]{#1}
\maketitle
\begin{center}
    \vspace{-12pt}
   \includegraphics[width=0.9\linewidth]{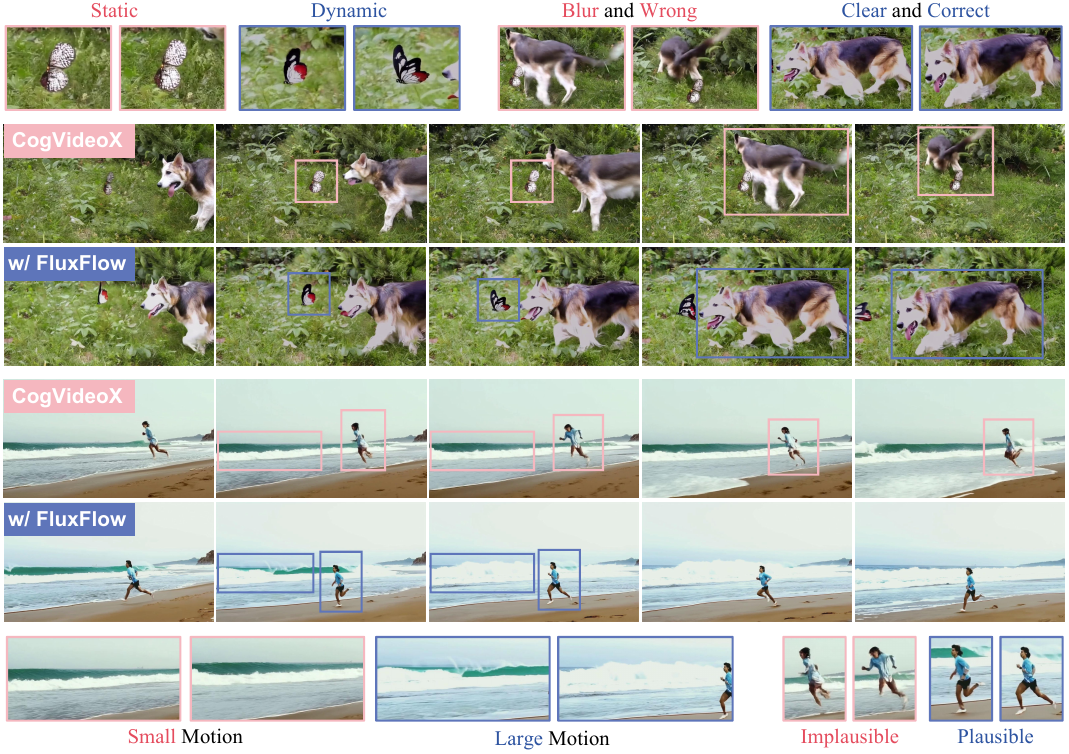}
   \vspace{-0.5em}
    \captionof{figure}{\label{fig:teaser}
    \textsc{FluxFlow} improves the temporal quality of video generators. Captions: \textit{(\textbf{Top}) A dog chasing a butterfly in a garden, with the butterfly flying in random directions. (\textbf{Bottom}) A person is running along a beach with waves crashing in the background.}}
\end{center}
}]

\begin{abstract}
Temporal quality is a critical aspect of video generation, as it ensures consistent motion and realistic dynamics across frames. However, achieving high temporal coherence and diversity remains challenging.  In this work, we explore temporal augmentation in video generation for the first time, and introduce \textsc{\textbf{FluxFlow}} for initial investigation, a strategy designed to enhance temporal quality. Operating at the data level, \textsc{FluxFlow} applies controlled temporal perturbations without requiring architectural modifications. Extensive experiments on UCF-101 and VBench benchmarks demonstrate that \textsc{FluxFlow} significantly improves temporal coherence and diversity across various video generation models, including U-Net, DiT, and AR-based architectures, while preserving spatial fidelity. These findings highlight the potential of temporal augmentation as a simple yet effective approach to advancing video generation quality.

\end{abstract}    
\section{Introduction}
\label{sec:intro}

\makeatletter
\renewcommand\subsubsection{\@startsection{subsubsection}{3}{\z@}%
                                     {-3.25ex\@plus -1ex \@minus .2ex}%
                                     {-1em}%
                                     {\normalfont\normalsize\bfseries}}
\makeatother

The pursuit of photorealistic video generation faces a critical dilemma: while spatial synthesis (\textit{e.g.}, SD-series \cite{rombach2022high, esser2024scaling}, AR-based \cite{wu2025lightgen, li2024autoregressive}) has achieved remarkable fidelity, ensuring temporal quality remains an elusive target. Modern video generators, whether diffusion \cite{kong2024hunyuanvideo, zheng2024open, li2025minimax, yang2024cogvideox, chen2024omnicreator} or autoregressive \cite{deng2024autoregressive, hong2022cogvideo, wang2024emu3}, frequently produce sequences plagued by temporal artifacts, \textit{e.g.}, flickering textures, discontinuous motion trajectories, or repetitive dynamics, exposing their inability to model temporal relationships robustly (see Figure~\ref{fig:teaser}).


These artifacts stem from a fundamental limitation: despite leveraging large-scale datasets, current models often rely on simplified temporal patterns in the training data (\textit{e.g.}, fixed walking directions or repetitive frame transitions) rather than learning diverse and plausible temporal dynamics. This issue is further exacerbated by the lack of explicit temporal augmentation during training, leaving models prone to overfitting to spurious temporal correlations (\textit{e.g.}, ``frame \#5 must follow \#4'') rather than generalizing across diverse motion scenarios. 

Unlike static images, videos inherently require models to reason about dynamic state transitions rather than isolated frames. While spatial augmentations (\textit{e.g.}, cropping, flipping, or color jittering) \cite{yun2019cutmix, zhang2017mixup} have proven effective for improving spatial fidelity in visual generation, they fail to address the temporal dimension, making them inadequate for video generation. As a result, video generation models often exhibit two key issues (as shown in Figure~\ref{fig:teaser}):
\ding{182} \textbf{\textit{Temporal inconsistency}}: Flickering textures or abrupt transitions between frames, indicating poor temporal coherence (see Figure~\ref{fig:optical}).
\ding{183} \textbf{\textit{Similar temporal patterns}}: Over-reliance on simplified temporal correlations leads to limited temporal diversity, where generated videos struggle to distinguish between distinct dynamics, such as fast and slow motion, even with explicit prompts (see Figure~\ref{fig:tsne}).
Addressing these challenges requires balancing spatial realism—textures, lighting, and object shapes—with temporal plausibility—coherent and diverse transitions. While modern architectures leverage large-scale image priors for spatial realism \cite{rombach2022high, esser2024scaling, sun2024autoregressive}, they struggle with complex temporal relationships, relying heavily on architectural modifications \cite{ho2022video, blattmann2023stable, guo2023animatediff} or constraint engineering \cite{wang2023lavie, chen2024videocrafter2, hong2022cogvideo}. However, data-level augmentation, proven effective in video understanding \cite{huang2025sefar, xing2023svformer, zou2023learning, chen2024finecliper}, remains underexplored, highlighting the untapped potential of temporal data augmentation for improving video generation.

To make an initial exploration of this issue, in this paper, we propose \textsc{\textbf{FluxFlow}}, a data augmentation strategy that injects controlled temporal perturbations into video generation training. Inspired by human cognition—where we infer missing frames or reorder events—\textsc{FluxFlow} operates on a simple principle: \textbf{\textit{disrupting fixed temporal order to force the model to learn disentangled motion/optical flow dynamics}}.
Specifically, \textsc{FluxFlow} introduces two levels of temporal perturbations for investigation:
\begin{itemize}[leftmargin=*]
    \item[\ding{238}] \textbf{Frame-Level}: Randomly shuffle individual frames to disrupt fixed temporal order, encouraging the model to infer plausible temporal relationships.
    \item[\ding{238}] \textbf{Block-Level}: Reorder contiguous-frame blocks to simulate realistic temporal disruptions while preserving coarse motion patterns.
\end{itemize}

\begin{figure}[t]
    \centering
   \includegraphics[width=1\linewidth]{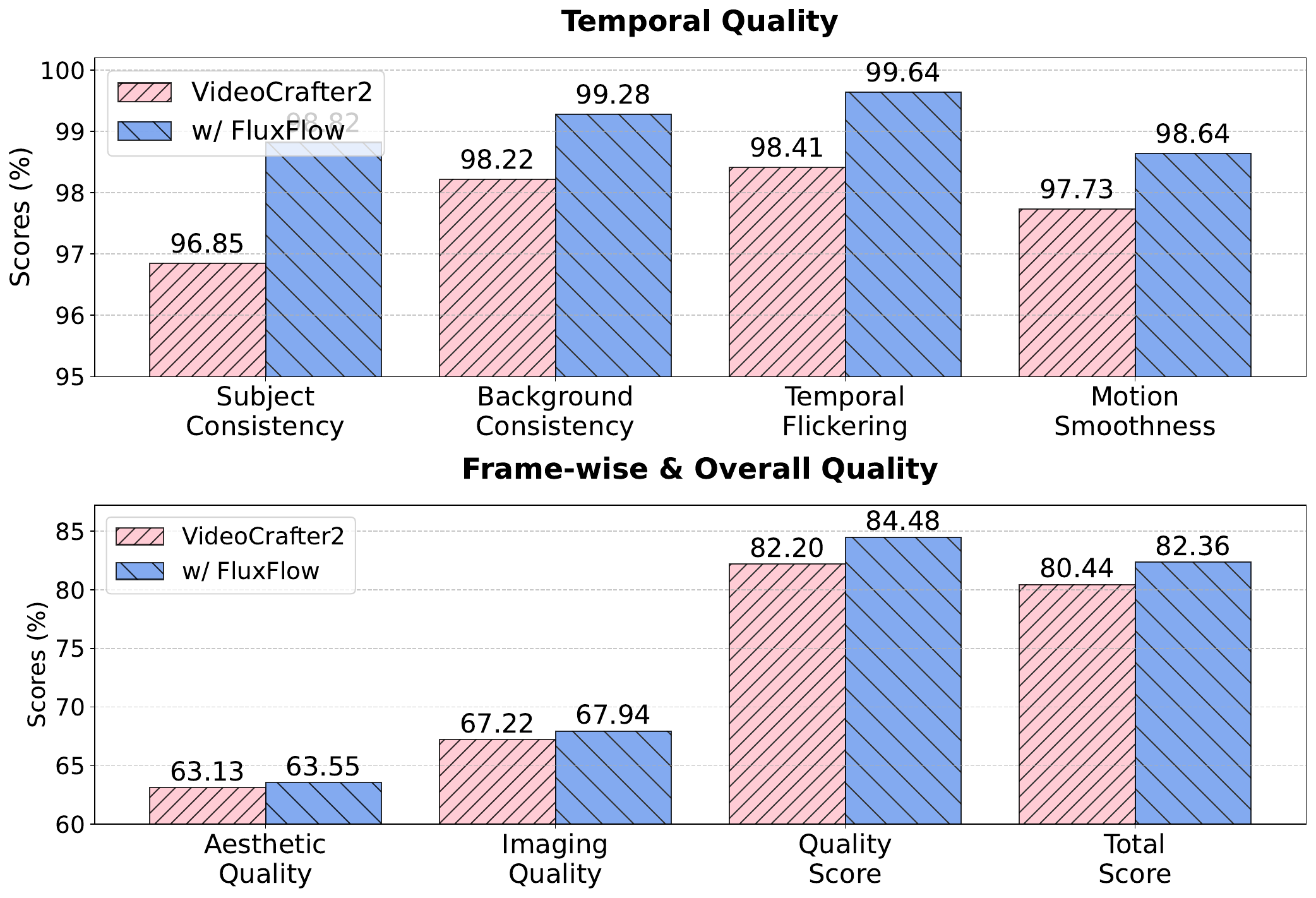}
   \vspace{-1.8em}
    \captionof{figure}{\label{fig:intro_vbench}
    Comparison of VideoCrafter2 with \textsc{FluxFlow} using VBench metrics for Temporal Quality (\textit{\textbf{Top}}) and Frame-wise and Overall Quality (\textit{\textbf{Bottom}}). \textsc{FluxFlow} significantly enhances the temporal quality of generated videos while maintaining or even improving frame-wise and overall quality.}
    \vspace{-1.2em}
\end{figure}

By training on disordered sequences, the generator learns to recover plausible trajectories, effectively regularizing temporal entropy. \textsc{FluxFlow} bridges the gap between discriminative and generative temporal augmentation, offering a plug-and-play enhancement solution for temporally plausible video generation while improving overall quality (see Figure~\ref{fig:teaser} and~\ref{fig:intro_vbench}). Unlike existing methods that introduce architectural changes or rely on post-processing, \textsc{FluxFlow} operates directly at the data level, introducing controlled temporal perturbations during training.
To summarize, our contributions are as follows:
\begin{itemize}[leftmargin=*]
    \item We introduce \textbf{\textsc{FluxFlow}}, the first dedicated temporal augmentation strategy for video generation, which introduces controlled temporal perturbations without requiring architectural modifications.
    \item We identify and formalize the challenge of temporal brittleness in video generation, highlighting the lack of explicit temporal augmentation in existing methods and demonstrating the potential of temporal augmentation as a simple yet viable solution.
    \item Extensive experiments across UCF-101 \cite{soomro2012ucf101} and VBench \cite{huang2023vbench} benchmarks on diverse video generators (U-Net \cite{chen2024videocrafter2}, DiT \cite{yang2024cogvideox}, and AR-based \cite{deng2024autoregressive}) demonstrate that \textsc{FluxFlow} enhances temporal coherence without compromising spatial fidelity.
\end{itemize}

\vspace{0.4em}
\noindent\textit{We hope this work inspires broader explorations of temporal augmentation strategies in video generation and beyond.}

\begin{figure*}[t]
    \centering
    \vspace{-0.4em}
   \includegraphics[width=1\linewidth]{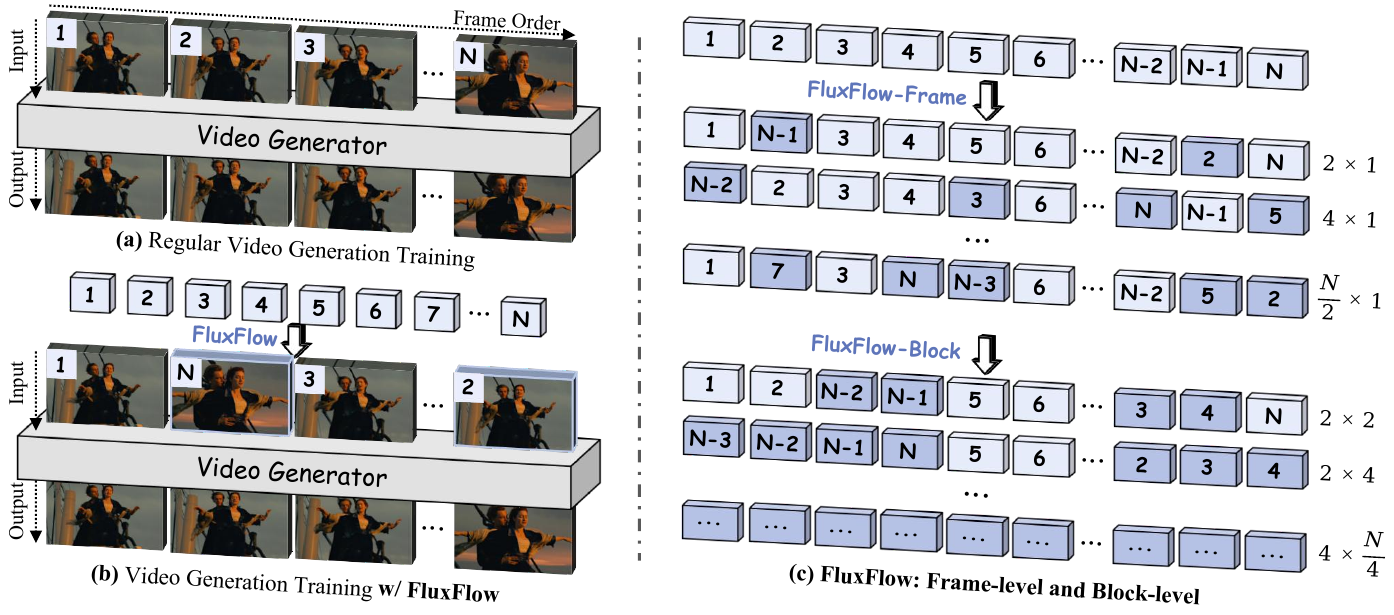}
   \vspace{-2em}
    \captionof{figure}{\label{fig:pipeline}
    Overview of \textsc{FluxFlow}. (\textit{\textbf{a}}) Standard video generation trains on fixed frame orders, which may limit the model's ability to learn temporal dynamics. (\textit{\textbf{b}}) \textsc{FluxFlow} introduces controlled temporal perturbations during training as a plug-and-play augmentation strategy. (\textit{\textbf{c}}) This study explores \textsc{FluxFlow} at two levels: frame-level (top) and block-level (bottom). In frame-level, $\text{Num} \times 1$ denotes the number of individual frames shuffled. In block-level, $\text{Num1} \times \text{Num2}$ represents a block comprising $\text{Num2}$ consecutive frames.}
    \vspace{-1.2em}
\end{figure*}
\section{Related Work}
\label{sec:related}
\vspace{-0.4em}

\subsubsection*{Video Generation.}
Advancements in video generation span T2V \cite{ho2022video, guo2023animatediff, kong2024hunyuanvideo, chen2024videocrafter2, wang2023modelscope, li2025minimax, hong2022cogvideo} and I2V \cite{xing2023dynamicrafter, blattmann2023stable, lin2024open, kong2024hunyuanvideo}. T2V generates videos aligned with textual descriptions, while I2V focuses on temporally coherent output conditioned on images. Beyond per-frame quality, ensuring temporal quality remains a key challenge.

\vspace{-1.4em}
\subsubsection*{Temporal Refinement for Video Generation.}

Modern approaches to temporal refinement can be categorized into three main paradigms: (\textit{\textbf{i}}) \textit{Architecture-Centric Modeling:} Spatiotemporal transformers \cite{ho2022video, li2025minimax}, hybrid 3D convolutions \cite{blattmann2023stable}, and motion-decoupled architectures \cite{guo2023animatediff} improve long-range coherence but increase computational cost. (\textit{\textbf{ii}}) \textit{Physics-Informed Regularization:} Techniques like optical flow warping \cite{wang2023lavie}, surface normal prediction \cite{chen2024videocrafter2}, and motion codebooks \cite{hong2022cogvideo} ensure realistic motion through physical priors. (\textit{\textbf{iii}}) \textit{Training Dynamics Optimization:} Temporal contrastive loss \cite{zhao2023motiondirector}, curriculum frame sampling \cite{lin2024open}, and dynamic FPS sampling \cite{kong2024hunyuanvideo} enhance robustness and consistency. While these methods have advanced architectural designs and constraint engineering, they often overlook the potential of systematic temporal augmentation within video data itself. Our work addresses this gap by introducing simple yet effective temporal augmentation strategies, paving the way for improved temporal quality in video generation.


\vspace{-0.4em}
\section{Methodology}

\vspace{-0.3em}
\subsection{Preliminaries}
\vspace{-0.3em}

Modern video generation models fall into three main paradigms: U-Net-based, Diffusion Transformer (DiT)-based, and Autoregressive (AR)-based. This section provides an overview of (Latent) Diffusion Models for U-Net and DiT, and Next Token Prediction for AR-based methods.

\vspace{0.3em}
\noindent\textbf{Diffusion Models (DMs)}~\cite{ho2020denoising, sohl2015deep} are probabilistic generative frameworks that gradually corrupt data $\mathbf{x}_0\sim p_{\mathrm{data}}(\mathbf{x})$ into Gaussian noise $\mathbf{x}_T\sim\mathcal{N}(0,\mathbf{I})$ via a forward process, and subsequently learn to reverse this process through denoising. The forward process $q(\mathbf{x}_t|\mathbf{x}_0,t)$,  defined over $T$ timesteps, progressively adds noise to the original data $\mathbf{x}_0$ to obtain $\mathbf{x}_t$, leveraging a parameterization trick. Conversely, the reverse process  $p_\theta(\mathbf{x}_{t-1}|\mathbf{x}_t,t)$ denoises $x_{t}$ to recover $x_{t-1}$ using a denoising network $\epsilon_\theta\left(\mathbf{x}_t,t\right)$. The training objective is formulated as follows:
\setlength\abovedisplayskip{3.4pt}
\setlength\belowdisplayskip{3.4pt}
\begin{equation}
    \min_\theta\mathbb{E}_{t,\mathbf{x}\sim p_\text{data},\epsilon\sim\mathcal{N}(0,\mathbf{I})}\|\epsilon-\epsilon_\theta\left(\mathbf{x}_t; \mathbf{c},t\right)\|_2^2, \label{eq_dm}
\end{equation}
where $\epsilon$ represents the ground-truth noise, $\theta$ denotes the learnable parameters, and $\mathbf{c}$ is an optional conditioning input. Once trained, the model generates data $\mathbf{x}_0$ by iteratively denoising a random Gaussian noise $\mathbf{x}_T$.

\vspace{0.4em}
\noindent\textbf{Latent Diffusion Models (LDMs)}~\cite{rombach2022high, ho2022imagen} extend DMs by operating in a compact latent space, significantly improving computational efficiency. Instead of performing the diffusion process in the pixel space, LDMs encode the input video $\mathbf{x}\in\mathbb{R}^{L\times3\times H\times W}$ into a latent representation $\mathbf{z}=\mathcal{E}(\mathbf{x})$ using an autoencoder $\mathcal{E}$, where $\mathbf{z}\in\mathbb{R}^{L\times C\times h\times w}$. The diffusion process $\mathbf{z}_{t}=p(\mathbf{z}_{0},t)$ and the denoising process $\mathbf{z}_t=p_\theta(\mathbf{z}_{t-1},\mathbf{c},t)$ are then conducted in the latent space. The training objective is similar to DMs but applied to the latent representation:
\setlength\abovedisplayskip{3.4pt}
\setlength\belowdisplayskip{3.4pt}
\begin{equation}
    \mathbb{E}_{t,\mathbf{x}\sim p_\text{data},\epsilon\sim\mathcal{N}(0,\mathbf{I})}\|\epsilon-\epsilon_\theta\left(\mathcal{E}(\mathbf{x}_t); \mathbf{c},t\right)\|_2^2, \label{eq_dm}
\end{equation}
Finally, the generated latent representation $\mathbf{z}$ is decoded back into the pixel space using the decoder $\mathcal{D}$, yielding the generated video $\hat{\mathbf{x}} = \mathcal{D}(\mathbf{z})$.

\vspace{0.4em}
\noindent\textbf{Next Token Prediction.} AR video generation can be formulated as next-token prediction, similar to language modeling. A video is converted into a sequence of discrete video tokens $\mathcal{T} = \{t_1, t_2, ...,t_n\}$ by tokenizers. Similar to LLMs, the next video token is predicted using past video tokens as context. Specifically, the training objective is to minimize the following negative log-likelihood (NLL) loss:
\begin{equation}
    \mathcal{L}_{NLL}=\sum_i-\log P(t_i|t_1,t_2,\ldots,t_{i-1};\Theta),
\end{equation}
where the conditional probability $P$ of the predicted next $t_i$ is modeled by a transformer decoder with parameters $\Theta$.

\vspace{-0.2em}
\subsection{\large\textsc{\textbf{FluxFlow}}} 
\vspace{-0.4em}
While spatial augmentations (\textit{e.g.}, flipping, cropping) are commonly employed to enhance spatial robustness, the temporal dimension remains under-regularized in video generation. To address this gap, we propose \textsc{FluxFlow}, a data-level temporal augmentation strategy that perturbs the temporal structure of video sequences during training. In this initial exploration, \textsc{FluxFlow} operates in two modes: Frame-level and Block-level Perturbations, each targeting distinct temporal scales, as demonstrated in Figure~\ref{fig:pipeline}.

\vspace{-1.2em}
\subsubsection*{Frame-Level Perturbations.}
\textsc{FluxFlow-Frame} introduces fine-grained disruptions by shuffling individual frames within a sequence. As shown in Figure~\ref{fig:pipeline}(c) (top), given a video sequence $V = \{F_1, F_2,\ldots,F_N\}$, we randomly shuffle a subset of frames, controlled by the perturbation ratio $\alpha$. Formally:
\begin{equation}
    V_{\mathrm{frame}}=\mathrm{Shuffle}(\{F_i\mid i\in\mathcal{S}\})+\{F_j\mid j\notin\mathcal{S}\},
\end{equation}
where $\mathcal{S}$ is a randomly selected subset of frames with $|\mathcal{S}|=\lfloor\alpha N\rfloor$. Frames outside $\mathcal{S}$ remain in their original positions, maintaining partial temporal consistency. This perturbation forces the model to reconstruct plausible temporal relationships, enhancing its ability to generalize beyond deterministic frame-to-frame dependencies.

\vspace{-1.2em}
\subsubsection*{Block-Level Perturbations.}
\textsc{FluxFlow-Block} operates at a coarser scale by reordering contiguous blocks of frames, as illustrated in Figure~\ref{fig:pipeline}(c) (bottom). The input sequence $V$ is divided into $M$ non-overlapping blocks of size $k$, such that:
\begin{equation}
    V_{\mathrm{block}}=\{B_1,B_2,\ldots,B_M\},
\end{equation}
where $B_m=\{F_{(m-1)k+1},\ldots,F_{mk}\}$. A subset $\mathcal{B}$ of these blocks is then randomly reordered with a probability $\beta$, producing:
\begin{equation}
    V_{\mathrm{block}}^{\mathrm{perturbed}}=\mathrm{Reorder}(\{B_m\mid m\in\mathcal{B}\})+\{B_n\mid n\notin\mathcal{B}\}.
\end{equation}
Block-level perturbations simulate realistic temporal disruptions, such as changes in motion speed or direction, while preserving coarse motion patterns.

\vspace{-1.3em}
\subsubsection*{Implementation.}
\textsc{FluxFlow} is implemented as a pre-processing strategy applied during training. Each perturbation (frame-level or block-level) is independently applied to evaluate its impact on temporal quality. Figure~\ref{fig:pipeline}(b) illustrates the combined training pipeline. A concrete illustration of the algorithm is given in the pseudocode below.

\vspace{-1em}
\begin{algorithm}[H]
\caption{\textsc{FluxFlow} Pseudocode}
\label{alg:fluxflow}
\begin{algorithmic}[1]
\Require Video $V = \{F_1, F_2, ...,F_N\}$, perturbation type $\mathrm{mode}\in\{\mathrm{frame},\mathrm{block}\}$, perturbation ratio $\alpha$ (for frame-level), block size $k$ and perturbation probability $\beta$ (for block-level)
\Ensure Perturbed sequence $V_{\text{FluxFlow}}$
\State \textbf{if} $\mathrm{mode}=\mathrm{frame}$: \Comment{\textcolor{gray}{Frame-Level Perturbations}}
\State \hspace{1em} Select subset $\mathcal{S}$ of frames with $|\mathcal{S}|=\lfloor\alpha N\rfloor$
\State \hspace{1em} Shuffle frames in $\mathcal{S}$ to obtain $V_{\text{FluxFlow}}$
\State \textbf{else if} $\mathrm{mode}=\mathrm{block}$: \Comment{\textcolor{gray}{Block-Level Perturbations}}
\State \hspace{1em} Divide $V$ into $M = \lfloor N/k\rfloor$ blocks $\{B_1,B_2,\ldots,B_M\}$
\State \hspace{1em} Select subset $\mathcal{B}$ of blocks with $|\mathcal{B}|=\lfloor\beta M\rfloor$
\State \hspace{1em} Reorder blocks in $\mathcal{B}$ to obtain $V_{\text{FluxFlow}}$
\State \textbf{end if}
\State \textbf{Output:} $V_{\text{FluxFlow}}$
\end{algorithmic}
\end{algorithm}
\vspace{-1.2em}

\begin{figure}[t]
    \centering
    \vspace{-0.4em}
   \includegraphics[width=1\linewidth]{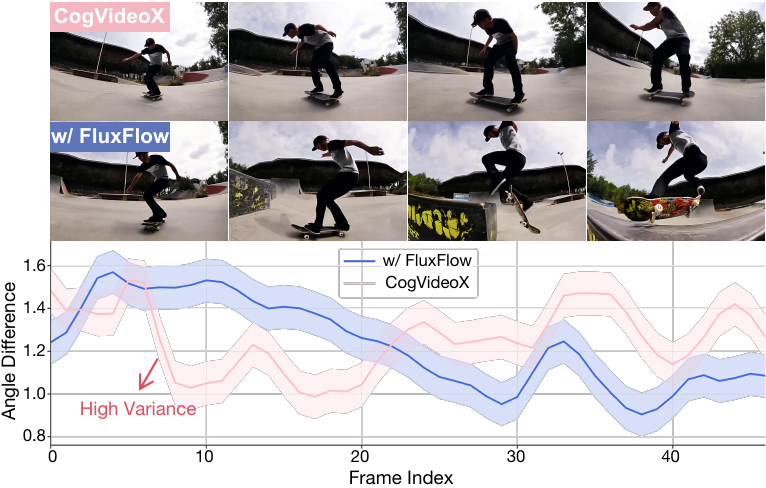}
   \vspace{-1.8em}
    \captionof{figure}{\label{fig:optical}
    Illustration of \textsc{FluxFlow} in enhancing temporal coherence. (\textit{\textbf{Top}}) Example frames from CogVideoX, without and with \textsc{FluxFlow}, showcasing larger motion dynamics in the latter. (\textbf{\textit{Bottom}}) Comparison of temporal angle differences across frames. \textsc{FluxFlow} achieves consistently lower angle differences, indicating improved temporal coherence over the base model. Caption: \textit{A skateboarder performing tricks in a skatepark, with fast-paced movements and dynamic camera angles.}}
    \vspace{-1.2em}
\end{figure}

\begin{figure}[t]
    \centering
    \vspace{-0.4em}
   \includegraphics[width=1\linewidth]{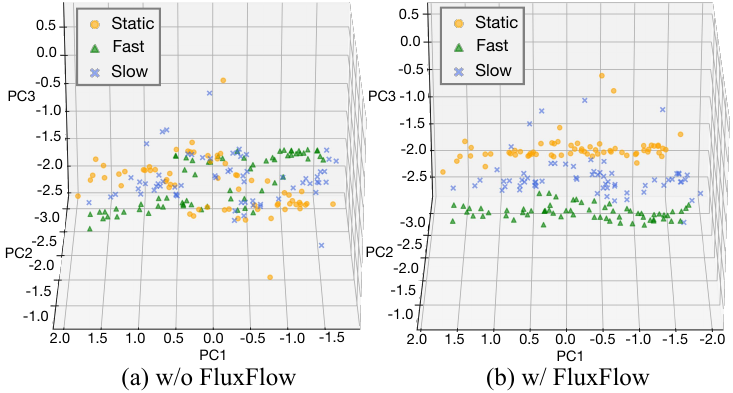}
   \vspace{-1.8em}
    \captionof{figure}{\label{fig:tsne}
    Illustration of \textsc{FluxFlow} in improving temporal feature diversity. (\textbf{\textit{a}}) Without \textsc{FluxFlow}, the model trained on fixed original frame sequences fails to distinguish features across different temporal paradigms. (\textit{\textbf{b}}) With \textsc{FluxFlow}, features are more distinctly separated, reflecting enhanced temporal representation.}
    \vspace{-1.2em}
\end{figure}

\vspace{-0.2em}
\subsection{What does model learn with \textsc{\textbf{FluxFlow}}?}
\label{sec:3.3}
\vspace{-0.4em}

To better understand the impact of \textsc{FluxFlow} on the model’s temporal learning capabilities, we evaluate its effect on temporal coherence and temporal diversity. For this purpose, we select three groups of text prompts with varying temporal dynamics: \textbf{static}, \textbf{slow}, and \textbf{fast} (details can be found in \textbf{Appendix}~\S\ref{app_setting}). Our observations are as follows:

\vspace{-1.2em}
\subsubsection*{Obs.\ding{182} \textsc{FluxFlow} enhances temporal coherence.} As shown in Figure~\ref{fig:optical}, we analyze videos generated from one of the ``fast" prompts. Videos generated without \textsc{FluxFlow} exhibit abrupt and unstable temporal changes, reflecting inconsistent motion dynamics. In contrast, the videos generated with \textsc{FluxFlow} demonstrate significantly larger and smoother motion dynamics. Quantitative analysis of angular differences further supports this observation. By comparing angular differences between consecutive frames, we observe that the base model produces high variance in these differences, reflecting erratic temporal transitions. In comparison, \textsc{FluxFlow} achieves consistently lower angular differences, indicating its ability to stabilize temporal changes while maintaining the intended motion dynamics.

\vspace{-1.2em}
\subsubsection*{Obs.\ding{183} \textsc{FluxFlow} improves temporal diversity.} Figure~\ref{fig:tsne} demonstrates the generated videos' temporal feature representations. Without \textsc{FluxFlow} (Figure~\ref{fig:tsne}(a)), the features of videos generated from different temporal prompts (static, slow, and fast) are largely overlapped, indicating the model struggles to distinguish between distinct temporal paradigms. This lack of separation reflects the baseline model’s inability to capture diverse temporal dynamics. In contrast, with \textsc{FluxFlow} (Figure~\ref{fig:tsne}(b)), the temporal features are more distinctly separated across the three temporal paradigms, reflecting the model’s enhanced ability to represent diverse temporal patterns.

These findings highlight the critical role of \textsc{FluxFlow} in improving the temporal capabilities of baseline models, allowing them to generate temporally consistent and diverse videos that align more closely with the intended motion dynamics of the input prompts.

\begingroup
\setlength{\tabcolsep}{2pt}
\begin{table*}[t]
\vspace{-0.4em}
\renewcommand{\arraystretch}{1.16}
  \centering
  \caption{Evaluation of \textsc{FluxFlow-Frame}. ``$+$ Original" refers to training without \textsc{FluxFlow}, while ``$+$ $\text{Num}\times 1$" indicates the use of different \textsc{FluxFlow-Frame} strategies. We \colorbox{cyan!10}{\textbf{shade}} the best results and \underline{underline} the second-best results for each model.}
  \centering
  \vspace{-1em}
  \scriptsize
   \begin{tabular}{l cc cccccccccc}
     \hlineB{2.5}
     \multirow{2}{*}{\textbf{Method}} & \multicolumn{2}{c}{\textbf{UCF-101}} & \multicolumn{10}{c}{\textbf{VBench}}\\
     \cmidrule(lr){2-3} \cmidrule(lr){4-13}
     & \textbf{FVD$\downarrow$} & \textbf{IS$\uparrow$} & \textbf{Subject$\uparrow$} & \textbf{Back.$\uparrow$} & \textbf{Flicker$\uparrow$} & \textbf{Motion$\uparrow$} & \textbf{Dynamic$\uparrow$} & \textbf{Aesthetic$\uparrow$} & \textbf{Imaging$\uparrow$} & \textbf{Quality$\uparrow$} & \textbf{Semantic$\uparrow$} & \textbf{Total$\uparrow$}\\
     \hlineB{2}
     \rowcolor{lightgray!20} \multicolumn{13}{c}{{\textit{U-Net-based}: 16F$\times$320$\times$512}} \\
     VideoCrafter2~\cite{chen2024videocrafter2}  & 463.80 & 36.57 & 96.85 & 98.22 & 98.41 & 97.73 & 42.50 & 63.13 & 67.22 & 82.20 & \underline{73.42} & 80.44\\
     $+$ Original & $\text{468.32}_{\textcolor{red!60!brown}{\uparrow4.52}}$ & $\text{37.13}_{\textcolor{ForestGreen}{\uparrow0.56}}$ & $\text{97.02}_{\textcolor{ForestGreen}{\uparrow0.17}}$ & $\text{97.89}_{\textcolor{red!60!brown!60!brown}{\downarrow0.33}}$ & $\text{97.17}_{\textcolor{red!60!brown}{\downarrow1.24}}$ & $\text{97.78}_{\textcolor{ForestGreen}{\uparrow0.05}}$ & $\text{41.24}_{\textcolor{red!60!brown}{\downarrow1.26}}$ & \cellcolor{cyan!10}\textbf{$\text{63.87}_{\textcolor{ForestGreen}{\uparrow0.74}}$} & \cellcolor{cyan!10}\textbf{$\text{68.01}_{\textcolor{ForestGreen}{\uparrow0.79}}$} & $\text{81.81}_{\textcolor{red!60!brown}{\downarrow0.39}}$ & $\text{73.14}_{\textcolor{red!60!brown}{\downarrow0.28}}$ & $\text{80.08}_{\textcolor{red!60!brown}{\downarrow0.36}}$\\
     \hdashline
     $+$ $2\times1$  & \cellcolor{cyan!10}\textbf{$\text{444.59}_{\textcolor{ForestGreen}{\downarrow19.21}}$} & \underline{$\text{37.89}$}$_{\textcolor{ForestGreen}{\uparrow1.32}}$ & \cellcolor{cyan!10}\textbf{$\text{98.82}_{\textcolor{ForestGreen}{\uparrow1.97}}$} & \cellcolor{cyan!10}\textbf{$\text{99.28}_{\textcolor{ForestGreen}{\uparrow1.06}}$} & \cellcolor{cyan!10}\textbf{$\text{99.64}_{\textcolor{ForestGreen}{\uparrow1.23}}$} & $\text{98.63}_{\textcolor{ForestGreen}{\uparrow0.90}}$ & \underline{$\text{49.58}$}$_{\textcolor{ForestGreen}{\uparrow7.08}}$ & \underline{$\text{63.55}$}$_{\textcolor{ForestGreen}{\uparrow0.42}}$ & \underline{$\text{67.94}$}$_{\textcolor{ForestGreen}{\uparrow0.72}}$ & \cellcolor{cyan!10}\textbf{$\text{84.48}_{\textcolor{ForestGreen}{\uparrow2.28}}$} & \cellcolor{cyan!10}\textbf{$\text{73.89}_{\textcolor{ForestGreen}{\uparrow0.47}}$} & \cellcolor{cyan!10}\textbf{$\text{82.36}_{\textcolor{ForestGreen}{\uparrow1.92}}$} \\
     $+$ $4\times1$  & \underline{$\text{451.43}$}$_{\textcolor{ForestGreen}{\downarrow12.37}}$ & $\text{37.02}_{\textcolor{ForestGreen}{\uparrow0.45}}$ & $\text{97.90}_{\textcolor{ForestGreen}{\uparrow1.05}}$ & \underline{$\text{99.15}$}$_{\textcolor{ForestGreen}{\uparrow0.93}}$ & $\text{98.66}_{\textcolor{ForestGreen}{\uparrow0.25}}$ & \underline{$\text{98.66}$}$_{\textcolor{ForestGreen}{\uparrow0.93}}$ & \cellcolor{cyan!10}\textbf{$\text{50.00}_{\textcolor{ForestGreen}{\uparrow7.50}}$} & $\text{61.74}_{\textcolor{red!60!brown}{\downarrow1.39}}$ & $\text{65.76}_{\textcolor{red!60!brown}{\downarrow1.46}}$ & \underline{$\text{83.31}$}$_{\textcolor{ForestGreen}{\uparrow1.11}}$ & $\text{73.39}$$_{\textcolor{red!60!brown}{\downarrow0.03}}$ & \underline{$\text{81.33}$}$_{\textcolor{ForestGreen}{\uparrow0.89}}$\\
     $+$ $8\times1$  & $\text{457.21}_{\textcolor{ForestGreen}{\downarrow6.59}}$ & \cellcolor{cyan!10}\textbf{$\text{37.92}_{\textcolor{ForestGreen}{\uparrow1.35}}$} & \underline{$\text{97.93}$}$_{\textcolor{ForestGreen}{\uparrow1.08}}$ & $\text{98.71}_{\textcolor{ForestGreen}{\uparrow0.49}}$ & \underline{$\text{98.69}$}$_{\textcolor{ForestGreen}{\uparrow0.28}}$ & \cellcolor{cyan!10}\textbf{$\text{98.92}_{\textcolor{ForestGreen}{\uparrow1.19}}$} & $\text{47.25}_{\textcolor{ForestGreen}{\uparrow4.75}}$ & $\text{60.97}_{\textcolor{red!60!brown}{\downarrow2.16}}$ & $\text{66.20}_{\textcolor{red!60!brown}{\downarrow1.02}}$ & $\text{83.11}_{\textcolor{ForestGreen}{\uparrow0.91}}$ & $\text{72.37}_{\textcolor{red!60!brown}{\downarrow1.05}}$ & $\text{80.96}_{\textcolor{ForestGreen}{\uparrow0.52}}$\\
     \hlineB{1.5}
     \rowcolor{lightgray!20} \multicolumn{13}{c}{{\textit{AR-based}: 33F$\times$480$\times$768}} \\
     NOVA~\cite{deng2024autoregressive}  & 428.12 & 38.44 & 94.71 & 94.81 & 96.38 & 96.34 & 54.35 & 54.52 & 66.21 & 78.96 & 76.57 & 78.48 \\
     $+$ Original & $\text{427.42}_{\textcolor{ForestGreen}{\downarrow0.70}}$ & \cellcolor{cyan!10}\textbf{$\text{39.49}_{\textcolor{ForestGreen}{\uparrow1.05}}$} & $\text{95.12}_{\textcolor{ForestGreen}{\uparrow0.41}}$ & $\text{94.54}_{\textcolor{red!60!brown}{\downarrow0.27}}$ & $\text{95.88}_{\textcolor{red!60!brown}{\downarrow0.50}}$ & $\text{96.45}_{\textcolor{ForestGreen}{\uparrow0.11}}$ & $\text{52.23}_{\textcolor{red!60!brown}{\downarrow2.12}}$ & \underline{$\text{54.89}$}$_{\textcolor{ForestGreen}{\uparrow0.37}}$ & \cellcolor{cyan!10}\textbf{$\text{67.04}_{\textcolor{ForestGreen}{\uparrow0.83}}$} & $\text{78.84}_{\textcolor{red!60!brown}{\downarrow0.12}}$ & \cellcolor{cyan!10}\textbf{$\text{76.87}_{\textcolor{ForestGreen}{\uparrow0.30}}$} & $\text{78.37}_{\textcolor{red!60!brown}{\downarrow0.11}}$ \\
     \hdashline
     $+$ $2\times1$ & \underline{$\text{420.17}$}$_{\textcolor{ForestGreen}{\downarrow7.95}}$ & $\text{38.71}_{\textcolor{ForestGreen}{\uparrow0.27}}$ & \underline{$\text{96.18}$}$_{\textcolor{ForestGreen}{\uparrow1.47}}$ & \underline{$\text{95.56}$}$_{\textcolor{ForestGreen}{\uparrow0.75}}$ & $\text{96.87}$$_{\textcolor{ForestGreen}{\uparrow0.49}}$ & \underline{$\text{97.40}$}$_{\textcolor{ForestGreen}{\uparrow1.06}}$ & \cellcolor{cyan!10}\textbf{$\text{58.64}$$_{\textcolor{ForestGreen}{\uparrow4.29}}$} & $\text{54.22}_{\textcolor{red!60!brown}{\downarrow0.30}}$ & \underline{$\text{66.86}$}$_{\textcolor{ForestGreen}{\uparrow0.65}}$ & \underline{$\text{80.52}$}$_{\textcolor{ForestGreen}{\uparrow1.56}}$ & $\text{76.11}$$_{\textcolor{red!60!brown}{\downarrow0.46}}$ & \underline{$\text{79.64}$}$_{\textcolor{ForestGreen}{\uparrow1.16}}$ \\
     $+$ $4\times1$ & \cellcolor{cyan!10}\textbf{$\text{413.45}_{\textcolor{ForestGreen}{\downarrow14.67}}$} & $\text{39.31}_{\textcolor{ForestGreen}{\uparrow0.87}}$ & \cellcolor{cyan!10}\textbf{$\text{96.76}_{\textcolor{ForestGreen}{\uparrow2.05}}$} & \cellcolor{cyan!10}\textbf{$\text{96.24}$$_{\textcolor{ForestGreen}{\uparrow1.43}}$} & \cellcolor{cyan!10}\textbf{$\text{97.45}$$_{\textcolor{ForestGreen}{\uparrow1.07}}$} & $\text{97.21}$$_{\textcolor{ForestGreen}{\uparrow0.87}}$ & \underline{$\text{57.88}$}$_{\textcolor{ForestGreen}{\uparrow3.53}}$ & \cellcolor{cyan!10}\textbf{$\text{54.96}_{\textcolor{ForestGreen}{\uparrow0.44}}$} & $\text{66.50}_{\textcolor{ForestGreen}{\uparrow0.29}}$ & \cellcolor{cyan!10}\textbf{$\text{80.91}$$_{\textcolor{ForestGreen}{\uparrow1.95}}$} & \underline{$\text{76.84}$}$_{\textcolor{ForestGreen}{\uparrow0.27}}$ & \cellcolor{cyan!10}\textbf{$\text{80.10}$$_{\textcolor{ForestGreen}{\uparrow1.62}}$} \\
     $+$ $16\times1$ & $\text{423.09}_{\textcolor{ForestGreen}{\downarrow5.03}}$ & \underline{$\text{39.24}$}$_{\textcolor{ForestGreen}{\uparrow0.89}}$ & $\text{95.24}_{\textcolor{ForestGreen}{\uparrow0.53}}$ & $\text{94.57}$$_{\textcolor{red!60!brown}{\downarrow0.24}}$ & \underline{$\text{97.12}$}$_{\textcolor{ForestGreen}{\uparrow0.74}}$ & \cellcolor{cyan!10}\textbf{$\text{97.52}$$_{\textcolor{ForestGreen}{\uparrow1.18}}$} & $\text{56.54}$$_{\textcolor{ForestGreen}{\uparrow2.20}}$ & $\text{54.18}_{\textcolor{red!60!brown}{\downarrow0.34}}$ & $\text{65.69}_{\textcolor{red!60!brown}{\downarrow0.52}}$ & $\text{79.97}$$_{\textcolor{ForestGreen}{\uparrow1.01}}$ & $\text{75.28}$$_{\textcolor{red!60!brown}{\downarrow1.29}}$ & $\text{79.03}$$_{\textcolor{ForestGreen}{\uparrow0.55}}$ \\
     \hlineB{1.5}
     \rowcolor{lightgray!20} \multicolumn{13}{c}{{\textit{DiT-based}: 49F$\times$480$\times$720}} \\
     CogVideoX~\cite{yang2024cogvideox} & 347.59 & 44.32 &  96.78 & 96.63 & 98.89 & 97.73 & 59.86 & \underline{60.82} & 61.68 & 82.18 & 75.83 & 80.91\\
     $+$ Original  & $\text{349.34}_{\textcolor{red!60!brown}{\uparrow1.75}}$ & \underline{$\text{45.91}$}$_{\textcolor{ForestGreen}{\uparrow1.59}}$ & $\text{96.82}_{\textcolor{ForestGreen}{\uparrow0.04}}$ & $\text{95.34}_{\textcolor{red!60!brown}{\downarrow1.29}}$ & $\text{98.83}_{\textcolor{red!60!brown}{\downarrow0.06}}$ & $\text{97.31}_{\textcolor{red!60!brown}{\downarrow0.42}}$ & $\text{60.16}_{\textcolor{ForestGreen}{\uparrow0.30}}$ & $\text{58.52}_{\textcolor{red!60!brown}{\downarrow2.30}}$ & \cellcolor{cyan!10}\textbf{$\text{62.25}_{\textcolor{ForestGreen}{\uparrow0.57}}$} & $\text{81.43}_{\textcolor{red!60!brown}{\downarrow0.76}}$ & $\text{75.96}_{\textcolor{ForestGreen}{\uparrow0.13}}$ & $\text{80.34}_{\textcolor{red!60!brown}{\downarrow0.57}}$\\
     \hdashline
     $+$ $2\times1$  & \underline{$\text{343.23}$}$_{\textcolor{ForestGreen}{\downarrow4.36}}$ & $\text{44.12}_{\textcolor{red!60!brown}{\downarrow0.20}}$ & $\text{97.32}_{\textcolor{ForestGreen}{\uparrow0.54}}$ & \underline{$\text{97.15}$}$_{\textcolor{ForestGreen}{\uparrow0.52}}$ & \underline{$\text{99.14}$}$_{\textcolor{ForestGreen}{\uparrow0.25}}$ & \underline{$\text{98.20}$}$_{\textcolor{ForestGreen}{\uparrow0.47}}$ & \underline{$\text{61.26}$}$_{\textcolor{ForestGreen}{\uparrow1.40}}$ & $\text{60.74}_{\textcolor{red!60!brown}{\downarrow0.08}}$ & $\text{61.96}_{\textcolor{ForestGreen}{\uparrow0.28}}$ & \underline{$\text{82.88}$}$_{\textcolor{ForestGreen}{\uparrow0.70}}$ & \underline{$\text{75.98}$}$_{\textcolor{ForestGreen}{\uparrow0.15}}$ & \underline{$\text{81.50}$}$_{\textcolor{ForestGreen}{\uparrow0.59}}$\\
     $+$ $8\times1$  & \cellcolor{cyan!10}\textbf{$\text{329.41}_{\textcolor{ForestGreen}{\downarrow18.18}}$} & \cellcolor{cyan!10}\textbf{$\text{46.09}_{\textcolor{ForestGreen}{\uparrow1.77}}$} & \cellcolor{cyan!10}\textbf{$\text{98.35}_{\textcolor{ForestGreen}{\uparrow1.57}}$} & \cellcolor{cyan!10}\textbf{$\text{97.98}_{\textcolor{ForestGreen}{\uparrow1.35}}$} & \cellcolor{cyan!10}\textbf{$\text{99.62}_{\textcolor{ForestGreen}{\uparrow0.73}}$} & \cellcolor{cyan!10}\textbf{$\text{98.24}_{\textcolor{ForestGreen}{\uparrow0.51}}$} & $\text{61.14}_{\textcolor{ForestGreen}{\uparrow1.28}}$ & \cellcolor{cyan!10}\textbf{$\text{61.54}_{\textcolor{ForestGreen}{\uparrow0.72}}$} & \underline{$\text{62.02}$}$_{\textcolor{ForestGreen}{\uparrow0.34}}$ & \cellcolor{cyan!10}\textbf{$\text{83.58}_{\textcolor{ForestGreen}{\uparrow1.40}}$} & \cellcolor{cyan!10}\textbf{$\text{76.09}_{\textcolor{ForestGreen}{\uparrow0.26}}$} & \cellcolor{cyan!10}\textbf{$\text{82.08}_{\textcolor{ForestGreen}{\uparrow1.17}}$} \\
     $+$ $24\times1$ & $\text{345.19}_{\textcolor{ForestGreen}{\downarrow2.40}}$ & $\text{44.98}_{\textcolor{ForestGreen}{\uparrow0.66}}$ & \underline{$\text{98.04}$}$_{\textcolor{ForestGreen}{\uparrow1.26}}$ & $\text{97.09}_{\textcolor{ForestGreen}{\uparrow0.46}}$ & $\text{98.96}_{\textcolor{ForestGreen}{\uparrow0.07}}$ & $\text{98.11}_{\textcolor{ForestGreen}{\uparrow0.38}}$ & \cellcolor{cyan!10}\textbf{$\text{62.15}_{\textcolor{ForestGreen}{\uparrow2.29}}$} & $\text{59.82}_{\textcolor{red!60!brown}{\downarrow1.00}}$ & $\text{60.21}_{\textcolor{red!60!brown}{\downarrow1.47}}$ & $\text{82.53}_{\textcolor{ForestGreen}{\uparrow0.35}}$ & $\text{74.29}_{\textcolor{red!60!brown}{\downarrow1.54}}$ & $\text{80.88}_{\textcolor{red!60!brown}{\downarrow0.03}}$ \\
     \hlineB{2.5}
   \end{tabular}
  \label{tab:fluxflow_frame}
  \vspace{-0.1cm}
\end{table*} 
\endgroup

\begingroup
\setlength{\tabcolsep}{2pt}
\begin{table*}[t]
\vspace{-0.2em}
\renewcommand{\arraystretch}{1.16}
  \centering
  \caption{Evaluation of \textsc{FluxFlow-Block}. ``$+$ $\text{Num1}\times \text{Num2}$" indicates the use of different \textsc{FluxFlow-Block} strategies.}
  \centering
  \vspace{-1em}
  \scriptsize
   \begin{tabular}{l cc cccccccccc}
     \hlineB{2.5}
     \multirow{2}{*}{\textbf{Method}} & \multicolumn{2}{c}{\textbf{UCF-101}} & \multicolumn{10}{c}{\textbf{VBench}}\\
     \cmidrule(lr){2-3} \cmidrule(lr){4-13}
     & \textbf{FVD$\downarrow$} & \textbf{IS$\uparrow$} & \textbf{Subject$\uparrow$} & \textbf{Back.$\uparrow$} & \textbf{Flicker$\uparrow$} & \textbf{Motion$\uparrow$} & \textbf{Dynamic$\uparrow$} & \textbf{Aesthetic$\uparrow$} & \textbf{Imaging$\uparrow$} & \textbf{Quality$\uparrow$} & \textbf{Semantic$\uparrow$} & \textbf{Total$\uparrow$}\\
     \hlineB{2}
     \rowcolor{lightgray!20} \multicolumn{13}{c}{{\textit{U-Net-based}: 16F$\times$320$\times$512}} \\
     VideoCrafter2~\cite{chen2024videocrafter2} & 463.80 & 36.57 & 96.85 & 98.22 & 98.41 & 97.73 & 42.50 & 63.13 & 67.22 & 82.20 & 73.42 & 80.44\\
     $+$ Original  & $\text{468.32}_{\textcolor{red!60!brown}{\uparrow4.52}}$ & $\text{37.13}_{\textcolor{ForestGreen}{\uparrow0.56}}$ & $\text{97.02}_{\textcolor{ForestGreen}{\uparrow0.17}}$ & $\text{97.89}_{\textcolor{red!60!brown}{\downarrow0.33}}$ & $\text{97.17}_{\textcolor{red!60!brown}{\downarrow1.24}}$ & $\text{97.78}_{\textcolor{ForestGreen}{\uparrow0.05}}$ & $\text{41.24}_{\textcolor{red!60!brown}{\downarrow1.26}}$ & \cellcolor{cyan!10}\textbf{$\text{63.87}_{\textcolor{ForestGreen}{\uparrow0.74}}$} & \underline{$\text{68.01}$}$_{\textcolor{ForestGreen}{\uparrow0.79}}$ & $\text{81.81}_{\textcolor{red!60!brown}{\downarrow0.39}}$ & $\text{73.14}_{\textcolor{red!60!brown}{\downarrow0.28}}$ & $\text{80.08}_{\textcolor{red!60!brown}{\downarrow0.36}}$\\
     \hdashline
     $+$ $2\times2$  & \cellcolor{cyan!10}\textbf{$\text{449.30}_{\textcolor{ForestGreen}{\downarrow14.50}}$} & \underline{$\text{37.76}$}$_{\textcolor{ForestGreen}{\uparrow1.19}}$ & \cellcolor{cyan!10}\textbf{$\text{98.32}_{\textcolor{ForestGreen}{\uparrow1.47}}$} & \underline{$\text{98.72}$}$_{\textcolor{ForestGreen}{\uparrow0.50}}$ & \cellcolor{cyan!10}\textbf{$\text{99.27}_{\textcolor{ForestGreen}{\uparrow0.86}}$} & \underline{$\text{98.63}$}$_{\textcolor{ForestGreen}{\uparrow0.90}}$ & \cellcolor{cyan!10}\textbf{$\text{48.85}_{\textcolor{ForestGreen}{\uparrow6.35}}$} & \underline{$\text{63.84}$}$_{\textcolor{ForestGreen}{\uparrow0.71}}$ & \cellcolor{cyan!10}\textbf{$\text{68.17}_{\textcolor{ForestGreen}{\uparrow0.95}}$} & \cellcolor{cyan!10}\textbf{$\text{84.14}_{\textcolor{ForestGreen}{\uparrow1.94}}$} & \underline{$\text{73.45}$}$_{\textcolor{ForestGreen}{\uparrow0.03}}$ & \cellcolor{cyan!10}\textbf{$\text{82.00}_{\textcolor{ForestGreen}{\uparrow1.56}}$} \\
     $+$ $2\times4$  & \underline{$\text{457.39}$}$_{\textcolor{ForestGreen}{\downarrow14.41}}$ & \cellcolor{cyan!10}\textbf{$\text{37.86}_{\textcolor{ForestGreen}{\uparrow1.29}}$} & \underline{$\text{97.59}$}$_{\textcolor{ForestGreen}{\uparrow0.74}}$ & \cellcolor{cyan!10}\textbf{$\text{99.18}_{\textcolor{ForestGreen}{\uparrow0.96}}$} & $\text{98.88}_{\textcolor{ForestGreen}{\uparrow0.47}}$ & \cellcolor{cyan!10}\textbf{$\text{98.85}_{\textcolor{ForestGreen}{\uparrow1.12}}$} & $\text{47.24}_{\textcolor{ForestGreen}{\uparrow4.74}}$ & $\text{63.24}_{\textcolor{ForestGreen}{\uparrow0.11}}$ & $\text{67.64}_{\textcolor{ForestGreen}{\uparrow0.42}}$ & \underline{$\text{83.76}$}$_{\textcolor{ForestGreen}{\uparrow1.56}}$ & \cellcolor{cyan!10}\textbf{$\text{73.67}_{\textcolor{ForestGreen}{\uparrow0.25}}$} & \underline{$\text{81.74}$}$_{\textcolor{ForestGreen}{\uparrow1.30}}$\\
     $+$ $4\times4$  & $\text{460.31}_{\textcolor{ForestGreen}{\downarrow14.67}}$ & $\text{36.41}_{\textcolor{red!60!brown}{\downarrow0.16}}$ & $\text{97.23}_{\textcolor{ForestGreen}{\uparrow0.38}}$ & $\text{98.45}_{\textcolor{ForestGreen}{\uparrow0.23}}$ & \underline{$\text{98.92}$}$_{\textcolor{ForestGreen}{\uparrow0.51}}$ & \underline{$\text{98.63}$}$_{\textcolor{ForestGreen}{\uparrow0.90}}$ & \underline{$\text{47.90}$}$_{\textcolor{ForestGreen}{\uparrow5.40}}$ & $\text{62.86}_{\textcolor{red!60!brown}{\downarrow0.27}}$ & $\text{66.90}_{\textcolor{red!60!brown}{\downarrow0.32}}$ & $\text{83.32}_{\textcolor{ForestGreen}{\uparrow1.12}}$ & $\text{72.08}_{\textcolor{red!60!brown}{\downarrow1.34}}$ & $\text{81.08}_{\textcolor{ForestGreen}{\uparrow0.64}}$\\
     \hlineB{1.5}
     \rowcolor{lightgray!20} \multicolumn{13}{c}{{\textit{AR-based}: 33F$\times$480$\times$768}} \\
     NOVA~\cite{deng2024autoregressive}  & 428.12 & 38.44 & 94.71 & 94.81 & 96.38 & 96.34 & 54.35 & 54.52 & 66.21 & 78.96 & 76.57 & 78.48 \\
     $+$ Original  & $\text{427.42}_{\textcolor{ForestGreen}{\downarrow0.70}}$ & \cellcolor{cyan!10}\textbf{$\text{39.49}_{\textcolor{ForestGreen}{\uparrow1.05}}$} & $\text{95.12}_{\textcolor{ForestGreen}{\uparrow0.41}}$ & $\text{94.54}_{\textcolor{red!60!brown}{\downarrow0.27}}$ & $\text{95.88}_{\textcolor{red!60!brown}{\downarrow0.50}}$ & $\text{96.45}_{\textcolor{ForestGreen}{\uparrow0.11}}$ & $\text{52.23}_{\textcolor{red!60!brown}{\downarrow2.12}}$ & \underline{$\text{54.89}$}$_{\textcolor{ForestGreen}{\uparrow0.37}}$ & \cellcolor{cyan!10}\textbf{$\text{67.04}_{\textcolor{ForestGreen}{\uparrow0.83}}$} & $\text{78.84}_{\textcolor{red!60!brown}{\downarrow0.12}}$ & \underline{$\text{76.87}$}$_{\textcolor{ForestGreen}{\uparrow0.30}}$ & $\text{78.37}_{\textcolor{red!60!brown}{\downarrow0.11}}$ \\
     \hdashline
     $+$ $2\times2$  & \underline{$\text{423.19}$}$_{\textcolor{ForestGreen}{\downarrow4.93}}$ & $\text{39.12}_{\textcolor{ForestGreen}{\uparrow0.68}}$ & $\text{96.24}_{\textcolor{ForestGreen}{\uparrow1.53}}$ & $\text{95.48}_{\textcolor{ForestGreen}{\uparrow0.67}}$ & $\text{96.89}_{\textcolor{ForestGreen}{\uparrow0.51}}$ & \underline{$\text{97.05}$}$_{\textcolor{ForestGreen}{\uparrow0.71}}$ & $\text{56.53}_{\textcolor{ForestGreen}{\uparrow2.18}}$ & $\text{54.24}_{\textcolor{red!60!brown}{\downarrow0.28}}$ & $\text{66.54}_{\textcolor{ForestGreen}{\uparrow0.33}}$ & $\text{80.13}_{\textcolor{ForestGreen}{\uparrow1.17}}$ & $\text{76.22}_{\textcolor{red!60!brown}{\downarrow0.35}}$ & $\text{79.35}_{\textcolor{ForestGreen}{\uparrow0.87}}$ \\
     $+$ $4\times4$  & \cellcolor{cyan!10}\textbf{$\text{417.99}_{\textcolor{ForestGreen}{\downarrow10.13}}$} & $\text{39.14}_{\textcolor{ForestGreen}{\uparrow0.70}}$ & \cellcolor{cyan!10}\textbf{$\text{96.89}_{\textcolor{ForestGreen}{\uparrow2.18}}$} & \underline{$\text{95.68}$}$_{\textcolor{ForestGreen}{\uparrow0.87}}$ & \cellcolor{cyan!10}\textbf{$\text{97.21}_{\textcolor{ForestGreen}{\uparrow0.83}}$} & $\text{96.89}_{\textcolor{ForestGreen}{\uparrow0.55}}$ & \cellcolor{cyan!10}\textbf{$\text{57.12}_{\textcolor{ForestGreen}{\uparrow2.78}}$} & \cellcolor{cyan!10}\textbf{$\text{55.02}_{\textcolor{ForestGreen}{\uparrow0.50}}$} & $\text{66.29}_{\textcolor{ForestGreen}{\uparrow0.08}}$ & \underline{$\text{80.47}$}$_{\textcolor{ForestGreen}{\uparrow1.51}}$ & \cellcolor{cyan!10}\textbf{$\text{76.97}_{\textcolor{ForestGreen}{\uparrow0.75}}$} & \underline{$\text{79.76}$}$_{\textcolor{ForestGreen}{\uparrow1.28}}$ \\
     $+$ $4\times8+1$ & $\text{425.04}_{\textcolor{ForestGreen}{\downarrow3.08}}$ & \underline{$\text{39.38}$}$_{\textcolor{ForestGreen}{\uparrow0.94}}$ & \underline{$\text{96.53}$}$_{\textcolor{ForestGreen}{\uparrow1.82}}$ & \cellcolor{cyan!10}\textbf{$\text{95.92}_{\textcolor{ForestGreen}{\uparrow1.11}}$} & \underline{$\text{97.04}$}$_{\textcolor{ForestGreen}{\uparrow0.66}}$ & \cellcolor{cyan!10}\textbf{$\text{97.24}_{\textcolor{ForestGreen}{\uparrow0.90}}$} & \underline{$\text{56.86}$}$_{\textcolor{ForestGreen}{\uparrow2.34}}$ & $\text{54.74}_{\textcolor{ForestGreen}{\uparrow0.22}}$ & \underline{$\text{66.86}$}$_{\textcolor{ForestGreen}{\uparrow0.65}}$ & \cellcolor{cyan!10}\textbf{$\text{80.59}_{\textcolor{ForestGreen}{\uparrow1.63}}$} & $\text{76.75}_{\textcolor{ForestGreen}{\uparrow0.18}}$ & \cellcolor{cyan!10}\textbf{$\text{79.82}_{\textcolor{ForestGreen}{\uparrow1.34}}$} \\
     \hlineB{1.5}
     \rowcolor{lightgray!20} \multicolumn{13}{c}{{\textit{DiT-based}: 49F$\times$480$\times$720}} \\
     CogVideoX~\cite{yang2024cogvideox}  & 347.59 & 44.32 & 96.78 & 96.63 & 98.89 & 97.73 & 59.86 & \underline{60.82} & 61.68 & 82.18 & 75.83 & 80.91\\
     $+$ Original  & $\text{349.34}_{\textcolor{red!60!brown}{\uparrow1.75}}$ & \underline{$\text{45.91}$}$_{\textcolor{ForestGreen}{\uparrow1.59}}$ & $\text{96.82}_{\textcolor{ForestGreen}{\uparrow0.04}}$ & $\text{95.34}_{\textcolor{red!60!brown}{\downarrow1.29}}$ & $\text{98.83}_{\textcolor{red!60!brown}{\downarrow0.06}}$ & $\text{97.31}_{\textcolor{red!60!brown}{\downarrow0.42}}$ & $\text{60.16}_{\textcolor{ForestGreen}{\uparrow0.30}}$ & $\text{58.52}_{\textcolor{red!60!brown}{\downarrow2.30}}$ & \underline{$\text{62.25}$}$_{\textcolor{ForestGreen}{\uparrow0.57}}$ & $\text{81.43}_{\textcolor{red!60!brown}{\downarrow0.76}}$ & \underline{$\text{75.96}$}$_{\textcolor{ForestGreen}{\uparrow0.13}}$ & $\text{80.34}_{\textcolor{red!60!brown}{\downarrow0.57}}$ \\
     \hdashline
     $+$ $2\times2$  & \underline{$\text{341.78}$}$_{\textcolor{ForestGreen}{\downarrow5.81}}$ & $\text{45.65}_{\textcolor{ForestGreen}{\uparrow1.33}}$ & $\text{97.14}_{\textcolor{ForestGreen}{\uparrow0.36}}$ & \underline{$\text{97.25}$}$_{\textcolor{ForestGreen}{\uparrow0.62}}$ & \cellcolor{cyan!10}\textbf{$\text{99.21}_{\textcolor{ForestGreen}{\uparrow0.32}}$} & $\text{98.05}_{\textcolor{ForestGreen}{\uparrow0.32}}$ & \underline{$\text{61.36}$}$_{\textcolor{ForestGreen}{\uparrow1.50}}$ & $\text{59.32}_{\textcolor{red!60!brown}{\downarrow1.50}}$ & \cellcolor{cyan!10}\textbf{$\text{62.86}_{\textcolor{ForestGreen}{\uparrow1.18}}$} & $\text{82.74}_{\textcolor{ForestGreen}{\uparrow0.56}}$ & $\text{75.85}_{\textcolor{ForestGreen}{\uparrow0.02}}$ & $\text{81.36}_{\textcolor{ForestGreen}{\uparrow0.45}}$ \\
     $+$ $4\times8$  & \cellcolor{cyan!10}\textbf{$\text{336.27}_{\textcolor{ForestGreen}{\downarrow11.32}}$} & \cellcolor{cyan!10}\textbf{$\text{45.93}_{\textcolor{ForestGreen}{\uparrow1.61}}$} & \cellcolor{cyan!10}\textbf{$\text{98.38}_{\textcolor{ForestGreen}{\uparrow2.05}}$} & \cellcolor{cyan!10}\textbf{$\text{97.81}_{\textcolor{ForestGreen}{\uparrow1.18}}$} & $\text{99.04}_{\textcolor{ForestGreen}{\uparrow0.15}}$ & \underline{$\text{98.46}$}$_{\textcolor{ForestGreen}{\uparrow0.73}}$ & \cellcolor{cyan!10}\textbf{$\text{61.85}_{\textcolor{ForestGreen}{\uparrow1.99}}$} & \cellcolor{cyan!10}\textbf{$\text{61.02}_{\textcolor{ForestGreen}{\uparrow0.20}}$} & $\text{62.18}_{\textcolor{ForestGreen}{\uparrow0.50}}$ & \cellcolor{cyan!10}\textbf{$\text{83.42}_{\textcolor{ForestGreen}{\uparrow1.24}}$} & \cellcolor{cyan!10}\textbf{$\text{76.36}_{\textcolor{ForestGreen}{\uparrow0.53}}$} & \cellcolor{cyan!10}\textbf{$\text{82.01}_{\textcolor{ForestGreen}{\uparrow1.10}}$} \\
     $+$ $4\times12+1$  & $\text{345.98}_{\textcolor{ForestGreen}{\downarrow1.61}}$ & $\text{45.19}_{\textcolor{ForestGreen}{\uparrow0.87}}$ & \underline{$\text{97.47}$}$_{\textcolor{ForestGreen}{\uparrow0.69}}$ & $\text{97.04}_{\textcolor{ForestGreen}{\uparrow0.41}}$ & \underline{$\text{99.18}$}$_{\textcolor{ForestGreen}{\uparrow0.29}}$ & \cellcolor{cyan!10}\textbf{$\text{98.68}_{\textcolor{ForestGreen}{\uparrow0.95}}$} & $\text{61.19}_{\textcolor{ForestGreen}{\uparrow1.33}}$ & $\text{59.64}_{\textcolor{red!60!brown}{\downarrow1.18}}$ & $\text{61.98}_{\textcolor{ForestGreen}{\uparrow0.30}}$ & \underline{$\text{82.98}$}$_{\textcolor{ForestGreen}{\uparrow0.80}}$ & $\text{75.92}_{\textcolor{ForestGreen}{\uparrow0.09}}$ & \underline{$\text{81.57}$}$_{\textcolor{ForestGreen}{\uparrow0.66}}$ \\
     \hlineB{2.5}
   \end{tabular}
  \label{tab:fluxflow_block}
  \vspace{-0.3cm}
\end{table*} 
\endgroup

\begin{figure*}[t]
    \centering
    \vspace{-0.4em}
   \includegraphics[width=1\linewidth]{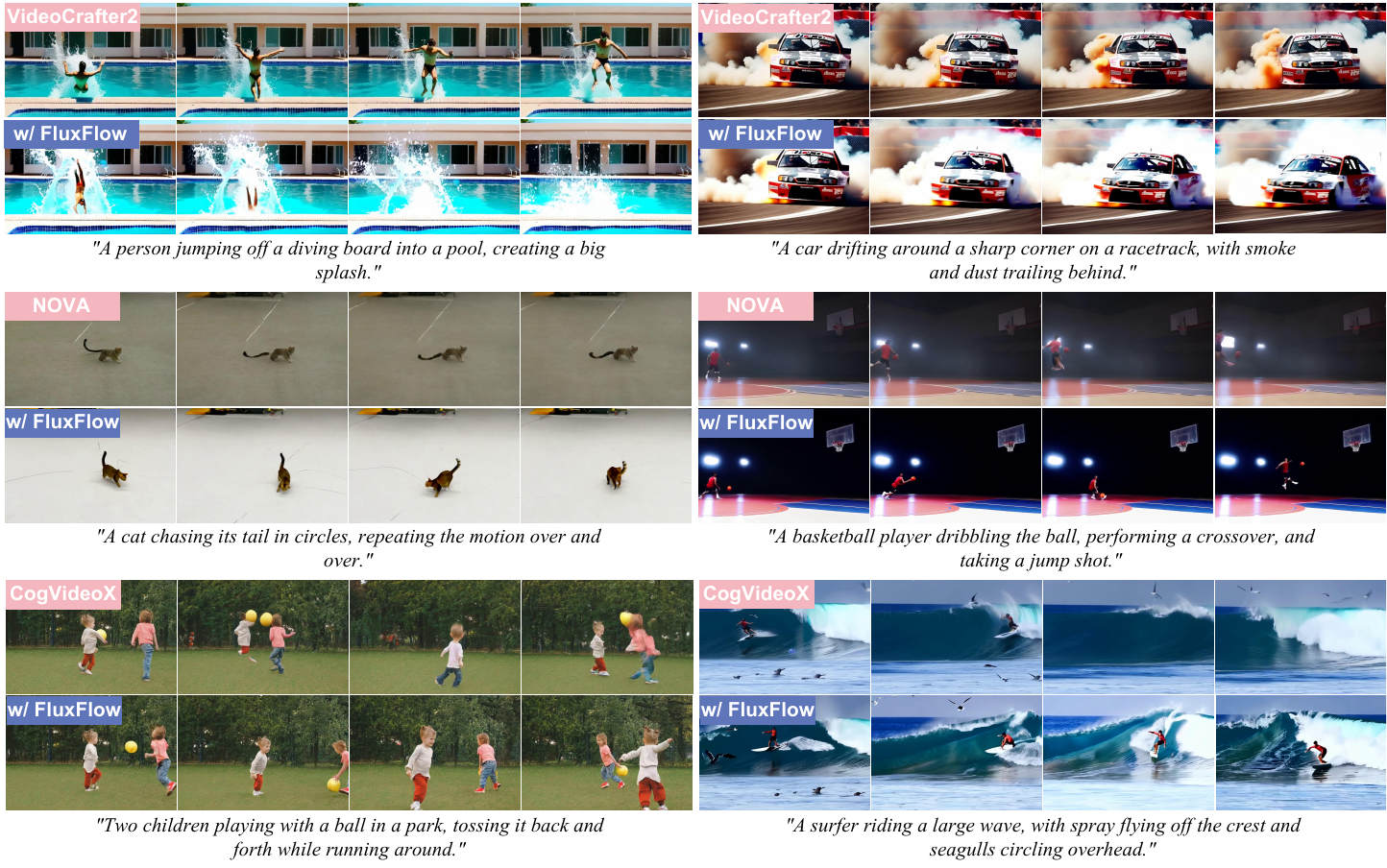}
   \vspace{-2em}
    \captionof{figure}{\label{fig:qualitative}
    Qualitative results of \textsc{FluxFlow} on VideoCrafter2 \cite{chen2024videocrafter2} (\textbf{\textit{Top}}), NOVA \cite{deng2024autoregressive} (\textbf{\textit{Middle}}), and CogVideoX \cite{yang2024cogvideox} (\textbf{\textit{Bottom}}).}
    \vspace{-0.8em}
\end{figure*}

\vspace{-0.4em}
\section{Experiment}
\label{sec:experi}
\vspace{-0.4em}

In this section, we conduct extensive experiments to answer the following research questions ($\mathcal{RQ}$):
\begin{enumerate}[start=1,label={\bfseries $\mathcal{RQ}$\arabic*:},leftmargin=3em]
\item Can \textsc{FluxFlow} improve temporal quality while maintaining spatial fidelity?
\item Does \textsc{FluxFlow} facilitate the learning of motion/optical flow dynamics?
\item Can \textsc{FluxFlow} maintain temporal quality in extra-term generation?
\item How sensitive is \textsc{FluxFlow} to its key hyperparameters?
\end{enumerate}

\vspace{-0.4em}
\subsection{Experimental Settings}
\vspace{-0.2em}

\subsubsection*{Base Models.} To comprehensively evaluate the effectiveness of \textsc{FluxFlow}, we apply it to three distinct video generation architectures: (\textit{i}) \textbf{U-Net-based}: VideoCrafter2 \cite{xing2023dynamicrafter}. (\textit{ii}) \textbf{AR-based}: NOVA-0.6B \cite{deng2024autoregressive}. (\textit{iii}) \textbf{DiT-based}: CogVideoX-2B \cite{yang2024cogvideox}.
To ensure fair and consistent comparisons, we fine-tune base models using \textsc{FluxFlow} as an additional training stage with one epoch on OpenVidHD-0.4M \cite{nan2024openvid}, following their default configurations (\textit{e.g.}, resolution, frame length). The results are compared with models trained under identical settings but without temporal augmentation (\textit{i.e.}, w/o \textsc{FluxFlow}). Notably, \textsc{FluxFlow} is model-agnostic and can be seamlessly integrated into the training pipeline of any video generation architecture.

\vspace{-1.3em}
\subsubsection*{Evaluations.} We evaluate \textsc{FluxFlow} on two widely-used benchmarks for video generation, focusing on both temporal coherence and overall video quality:
\begin{itemize}[leftmargin=*]
    \item \textbf{UCF-101} \cite{soomro2012ucf101}: A large-scale human action dataset containing $13,320$ videos across $101$ action classes. We utilize the following metrics: 
    \begin{itemize}[leftmargin=1.4em]
    \item[(\textit{i})] Fr\'{e}chet Video Distance (FVD) \cite{unterthiner2018towards} for temporal coherence and motion realism.
    \item[(\textit{ii})] Inception Score (IS) \cite{salimans2016improved} for frame-level quality and diversity. 
    \end{itemize}
    \item \textbf{VBench} \cite{huang2023vbench}: A comprehensive benchmark designed to evaluate video generation quality across $16$ dimensions. To specifically assess temporal and frame-level quality, we focus on the following key dimensions:
    \begin{itemize}[leftmargin=1.4em]
    \item[(\textit{i})] Temporal Quality: Subject Consistency, Background Consistency, Temporal Flickering, Motion Smoothness, and Dynamic Degree. \item[(\textit{ii})] Frame-Wise Quality: Aesthetic Quality and Imaging Quality. \item[(\textit{iii})] Overall Quality: Total Score, Quality Score, and Semantic Score. 
    \end{itemize}
\end{itemize}
These benchmarks and metrics provide a comprehensive evaluation, allowing us to rigorously assess the impact of \textsc{FluxFlow} on both temporal dynamics and spatial fidelity.

\vspace{-0.2em}
\subsection{Quality and Fidelity Enhancement (\texorpdfstring{$\mathcal{RQ}$}{RQ}1)}
\vspace{-0.1em}
We present the quantitative comparison of \textsc{FluxFlow-Frame} and \textsc{FluxFlow-Block} on VideoCrafter2 (VC2), NOVA, and CogVideoX (CVX) in Tab.~\ref{tab:fluxflow_frame} and~\ref{tab:fluxflow_block} and qualitative comparison in Fig.~\ref{fig:qualitative}. Each model is evaluated with three settings based on its default frame length. Specifically, \textsc{FluxFlow-Frame} is shown on VC2, NOVA, and CVX with $2\times1$, $4\times1$, and $8\times1$ in the qualitative comparisons, respectively. We give the following observations:

\vspace{-1.2em}
\subsubsection*{Obs.\ding{184} \textsc{FluxFlow} improves temporal quality with preserved spatial fidelity.} Both \textsc{FluxFlow-Frame} and \textsc{FluxFlow-Block} significantly improve temporal quality, as evidenced by the metrics in Tabs.~\ref{tab:fluxflow_frame},~\ref{tab:fluxflow_block} (\textit{i.e.}, FVD, Subject, Flicker, Motion, and Dynamic) and qualitative results in Fig.~\ref{fig:qualitative}. For instance, the motion of the drifting car in VC2, the cat chasing its tail in NOVA, and the surfer riding a wave in CVX become noticeably more fluid with \textsc{FluxFlow}. Importantly, these temporal improvements are achieved without sacrificing spatial fidelity, as evidenced by the sharp details of water splashes, smoke trails, and wave textures, along with spatial and overall fidelity metrics.

\vspace{-1.2em}
\subsubsection*{Obs.\ding{185} Optimal temporal perturbation strength is model-specific.} The ideal perturbation strength depends on the base model's default frame length. For example, in Tab.~\ref{tab:fluxflow_frame}, the 16-frame VC2 performs best with the $2\times1$ strategy, while the 49-frame CVX benefits most from $8\times1$. Excessive perturbation, however, may disrupt spatial consistency, highlighting the importance of selecting model-specific perturbation during training.

\vspace{-1.2em}
\subsubsection*{Obs.\ding{186} Frame-level perturbations outperform block-Level.} While both frame-level and block-level perturbations improve temporal quality, frame-level generally delivers better results. This can be attributed to their finer granularity, which allows for more precise temporal adjustments. In contrast, block-level perturbations may introduce excessive noise due to stronger spatiotemporal correlations within blocks, limiting their effectiveness. As a result, frame-level strategies yield smoother and more coherent motion transitions.

\subsection{User Study with Temporal Dynamics (\texorpdfstring{$\mathcal{RQ}$}{RQ}2)}
\label{sec:4.3}
To answer $\mathcal{RQ}$2, we first refer to Fig.~\ref{fig:optical}, which highlights \textsc{FluxFlow}'s ability to capture smooth and coherent optical flow changes, particularly in complex motion scenarios, and Fig.~\ref{fig:qualitative}, which demonstrates its superior motion realism and dynamics. Building on these findings, we further conduct a user study (Fig.~\ref{fig:user_study}) on $20$ video-pairs to evaluate subjective perceptions of motion quality across five dimensions: Motion Diversity, Motion Realism, Motion Smoothness, Temporal Coherence, and Optical Flow Consistency, using prompts of two types: \textit{\textbf{Action Speed}} (Fast \& Slow) and \textit{\textbf{Motion Pattern}} (Linear \& Nonlinear).
We observe that:

\vspace{-1.2em}
\subsubsection*{Obs.\ding{187} \textsc{FluxFlow} significantly facilitates temporal dynamics learning.} 
\begin{figure}[t]
    \centering
    \vspace{-0.4em}
   \includegraphics[width=1\linewidth]{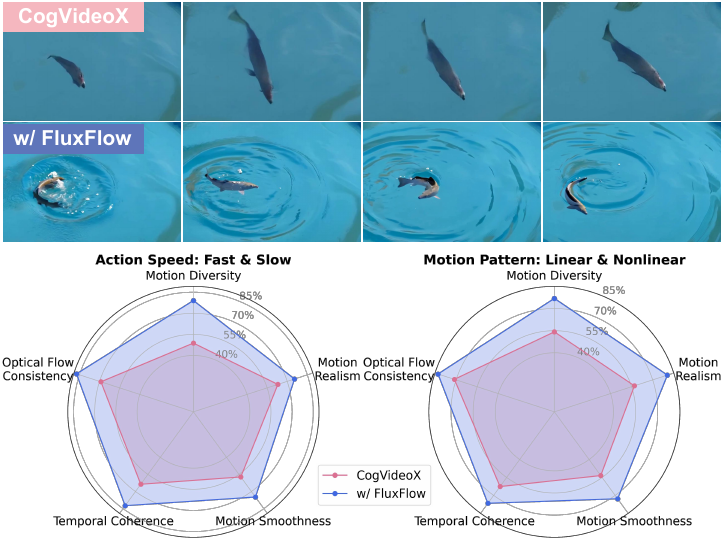}
   \vspace{-1.8em}
    \captionof{figure}{\label{fig:user_study}
    User study results comparing CVX and w/ \textsc{FluxFLow}. (\textbf{\textit{Top}}) Examples frames from a non-linear motion pattern, where \textsc{FluxFlow} demonstrates superior handling of complex trajectories. Caption: \textit{A fish swims in \textbf{circular loops} in a clear blue pond.} (\textbf{\textit{Bottom}}) User ratings across temporal dynamics evaluation criteria. For more details please refer to \textbf{Appendix}~\S\ref{app_setting}.}
    \vspace{-1em}
\end{figure}

As shown in Fig.~\ref{fig:user_study}, \textsc{FluxFlow} effectively disentangles and learns motion dynamics, excelling in complex trajectories and rapid temporal variations. Specifically, (\textit{\textbf{i}}) Motion Diversity: Broader and more varied motion trajectories, particularly in dynamic or nonlinear scenarios. (\textbf{\textit{ii}}) Optical Flow Consistency: Smoother and more coherent transitions, reducing abrupt changes and artifacts. (\textit{\textbf{iii}}) Motion Realism and Smoothness: More natural and fluid motion, especially in intricate and complex trajectories. (\textit{\textbf{iv}}) Temporal Coherence: Stable frame-to-frame dynamics without compromising other dimensions.

\subsection{Extra-term Temporal Quality (\texorpdfstring{$\mathcal{RQ}$}{RQ}3)}
To answer $\mathcal{RQ}$3 and evaluate whether \textsc{FluxFlow} can maintain temporal quality in extreme conditions, we specifically use the 16-frame VC2 to generate 128-frame videos, as shown in Fig.~\ref{fig:extra}. This allows us to verify whether \textsc{FluxFlow} can overcome the cumulative error and temporal instability challenges commonly observed in long-sequence generation. We give the following observations:

\begin{figure*}[t]
    \centering
    \vspace{-0.6em}
   \includegraphics[width=1\linewidth]{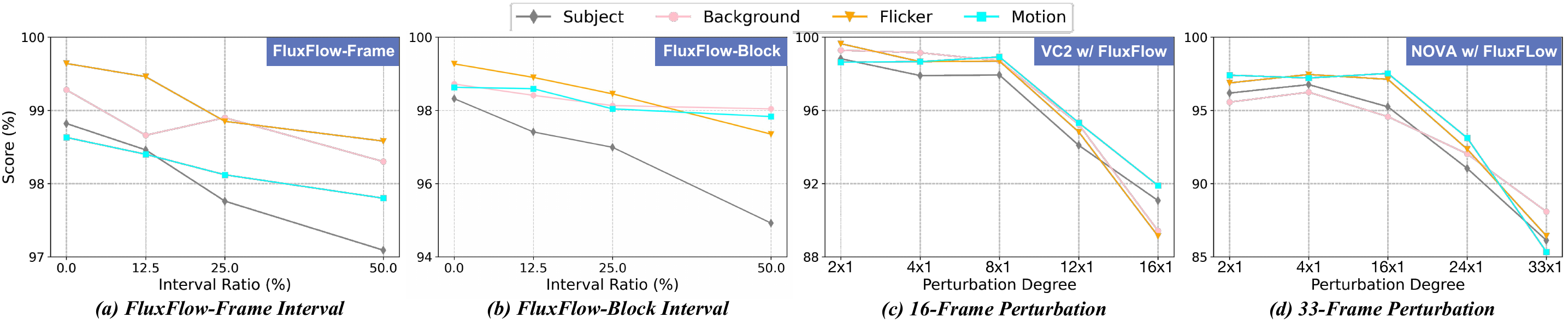}
   \vspace{-2em}
    \captionof{figure}{\label{fig:ablation}
    Ablation and sensitivity analysis on \textsc{FluxFlow} with VBench temporal metrics. (\textit{\textbf{a}}, \textit{\textbf{b}}) Impact of shuffle interval constraints on VC2 using $2\times1$ and $2\times2$ configurations. (\textbf{\textit{c}}, \textit{\textbf{d}}) Impact of perturbation degrees on 16-frame VC2 and 33-frame NOVA.}
    \vspace{-0.8em}
\end{figure*}

\vspace{-1.2em}
\subsubsection*{Obs.\ding{188} \textsc{FluxFlow} effectively preserves temporal quality under extreme conditions.}
\begin{figure}[t]
    \centering
    \vspace{-0.4em}
   \includegraphics[width=1\linewidth]{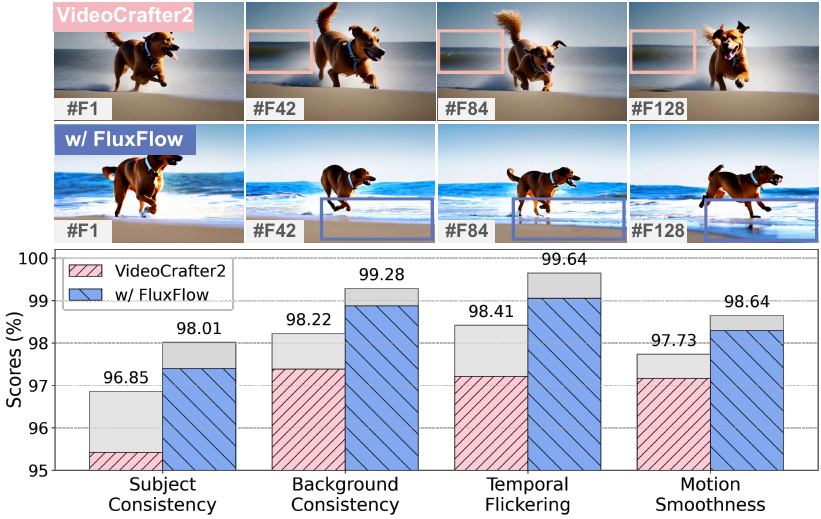}
   \vspace{-2em}
    \captionof{figure}{\label{fig:extra}
    Performance comparison under extra-term conditions. (\textit{\textbf{Top}}) Example frames from 16-frame VC2 generating 128-frame, without and with \textsc{FluxFlow}, showcasing dynamic background consistency in the latter. Caption: \textit{A dog running along a beach, splashing water as it moves through the waves.} (\textbf{\textit{Bottom}}) Comparison of temporal quality metrics on VBench, where the gray regions indicate the performance drop under extra-term scenarios.}
    \vspace{-1em}
\end{figure}

As shown in Fig.~\ref{fig:extra}, the qualitative comparison (top) demonstrates that \textsc{FluxFlow} maintains dynamic background consistency and generates smoother transitions, while the baseline (VC2) exhibits temporal artifacts, \textit{e.g.}, flickering and motion inconsistency. Quantitatively (bottom), the gray regions highlight score drops relative to the original 16-frame generation. \textsc{FluxFlow} significantly reduces these drops, achieving superior subject consistency, background consistency, temporal flickering, and motion smoothness scores, ensuring high temporal quality in extra-term scenarios.



\subsection{Ablation \& Sensitivity Analysis (\texorpdfstring{$\mathcal{RQ}$}{RQ}4)}

To better investigate the effectiveness of \textsc{FluxFlow}, we conduct two ablation studies to assess its sensitivity to shuffle interval constraints and perturbation degrees in Fig.~\ref{fig:ablation}: (\textit{\textbf{i}}) Inter-frame/block Interval, and (\textit{\textbf{ii}}) Perturbation Degree.

\vspace{-1.2em}
\subsubsection*{Inter-frame/block Interval Analysis.} We analyze the impact of shuffle interval constraints on frame-level (\textsc{FluxFlow-Frame}) and block-level (\textsc{FluxFlow-Block}). The shuffle interval defines the minimum distance between shuffled frames or blocks. For example, in a $2\times1$ frame-level shuffle with an interval of $8$ frames, any two shuffled frames must be separated by at least $8$ frames. As demonstrated in Fig.~\ref{fig:ablation}(a,b), ablations on VC2 using $2\times1$ and $2\times2$ shuffle configurations reveal that removing interval constraints (0.0\% interval ratio) achieves the best performance across all metrics. Larger constraints (\textit{e.g.}, 25\% or 50\%) lead to noticeable performance degradation. This suggests that allowing free shuffle without interval constraints enables the model to better leverage temporal information, supporting the hypothesis that excessive constraints reduce the diversity of temporal patterns learned by the model.

\vspace{-1.2em}
\subsubsection*{Perturbation Degree Analysis.} We further examine whether excessive perturbation would cause significant performance degradation. We performed frame-level ablation on 16-frame VC2 and 33-frame NOVA, as illustrated in Fig.~\ref{fig:ablation}(c,d). The results indicate that performance begins to decline significantly when the perturbation degree exceeds half of the total frames. This observation aligns with \textbf{Obs}\ding{185}, which highlights that perturbing more than half of the frames disrupts the model's ability to infer the correct temporal order due to insufficient contextual information.

\vspace{-0.3em}
\section{Conclusion}
\label{sec:conclu}
\vspace{-0.3em}

In this work, we propose \textsc{FluxFlow}, a pioneering temporal data augmentation strategy aimed at enhancing temporal quality in video generation. This initial exploration introduces two simple yet effective ways: frame-level (\textsc{FluxFlow-Frame}) and block-level (\textsc{FluxFlow-Block}). By addressing the limitations of existing methods that focus primarily on architectural designs and condition-informed constraints, \textsc{FluxFlow} bridges a critical gap in the field. Extensive experiments demonstrate that integrating \textsc{FluxFlow} significantly improves both temporal coherence and overall video fidelity. We believe \textsc{FluxFlow} sets a promising foundation for future research in temporal augmentation strategies, paving the way for more robust and temporally consistent video generation.

{
    \small
    \bibliographystyle{ieeenat_fullname}
    \bibliography{main}
}
\clearpage
\maketitlesupplementary
\appendix


\section{Detailed Analysis Settings}
\label{app_setting}

This section details the prompt information for the analysis in the main text.

\subsection{Temporal Diversity Analysis}
As shown in Figure~\ref{fig:tsne} in Section~\ref{sec:3.3}, we analyze temporal diversity within generated videos by evaluating three groups of text prompts with distinct temporal dynamics: \textbf{Static}, \textbf{Slow}, and \textbf{Fast}. These prompts are designed to capture varying levels of motion and temporal changes across the generated videos. Below, we provide the complete prompt details for each group:

\begin{figure}[h]
    \centering
    \vspace{-1em}
   \includegraphics[width=0.9\linewidth]{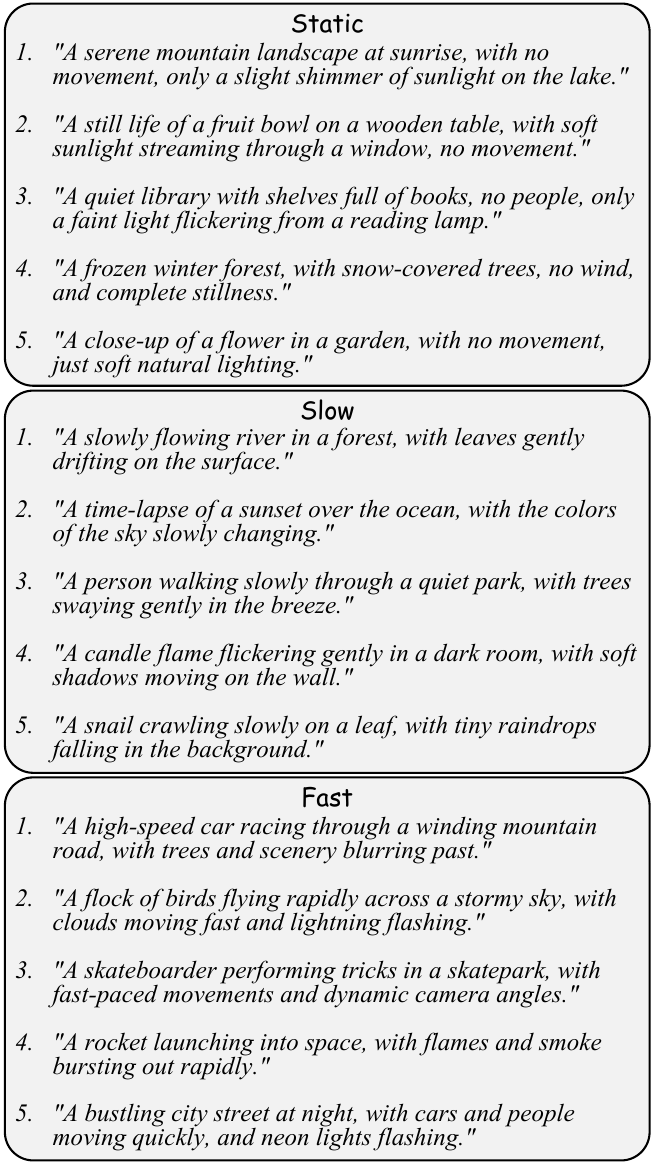}
    \vspace{-1em}
\end{figure}

These prompts ensure a comprehensive assessment of the model's ability to generate videos with diverse temporal characteristics.

\subsection{User Study with Temporal Dynamics}
In Section~\ref{sec:4.3}, we conduct user studies to evaluate the perceived quality of temporal dynamics in the generated videos. The study involves two key aspects: \textbf{Action Speed} (Fast \& Slow) and \textbf{Motion Pattern} (Linear \& Nonlinear). Below, we provide the full prompt details used for each category:
\begin{figure}[h]
    \centering
    \vspace{-1em}
   \includegraphics[width=0.9\linewidth]{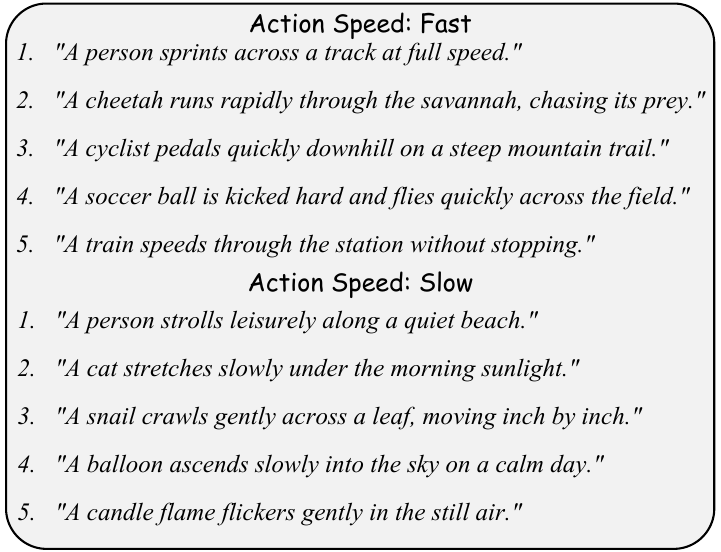}
    \vspace{-1em}
\end{figure}
\begin{figure}[h]
    \centering
    \vspace{-1em}
   \includegraphics[width=0.9\linewidth]{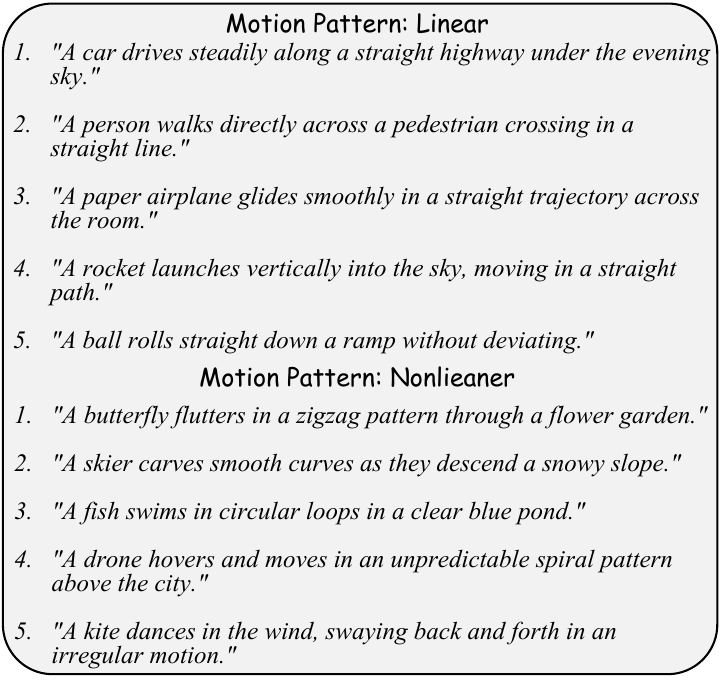}
    \vspace{-1em}
\end{figure}

\begin{figure*}[t]
    \centering
   \includegraphics[width=0.85\linewidth]{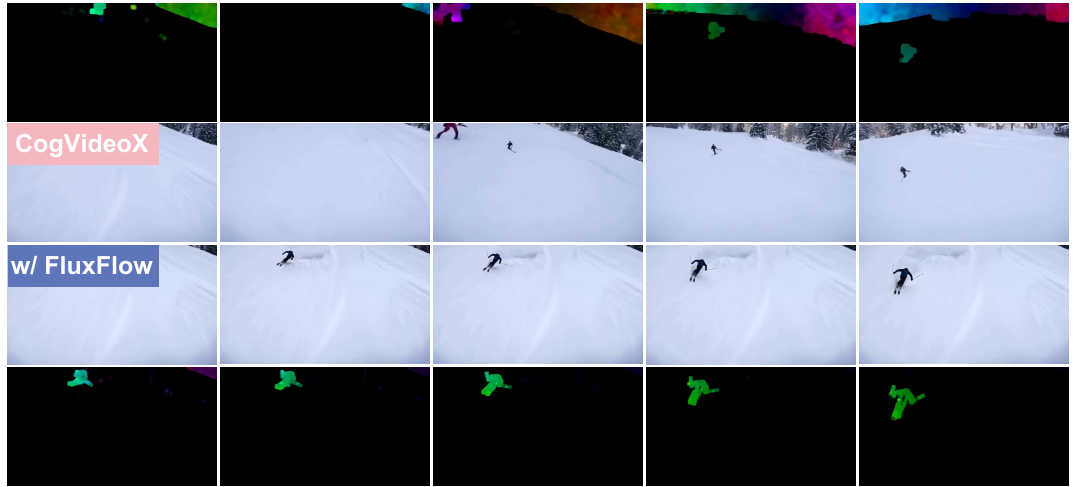}
   \vspace{-0.4em}
    \captionof{figure}{\label{fig:app_user}
    User study examples. Each video is provided with its optical flow to assess the Optical Flow Consistency. Caption: \textit{A skier carves smooth curves as they descend a snowy slope.}}
    \vspace{-0.8em}
\end{figure*}

Ten participants were asked to evaluate the generated videos from Motion Diversity, Motion Realism, Motion Smoothness, Temporal Coherence, and Optical Flow Consistency, scoring from $0$ to $5$ for each dimension. The results of this study provide valuable insights into the effectiveness of \textsc{FluxFlow} in capturing different temporal dynamics. Examples are shown in Figure~\ref{fig:app_user}.

\section{Limitations}
\label{app_details}
Deep learning \cite{chen2024gaussianvton, ma2024beyond, huang2024crest, yan2024urbanclip, wu2024fsc, 10613828, deng2024exploring, shao2020finegym, huang2024trusted, chen2024bovila, liu2024seeing, he2024efficient, liu2024mycloth} has revolutionized video generation by enabling models to learn complex spatiotemporal patterns from large-scale data. While our work introduces \textsc{FluxFlow} as a pioneering exploration of temporal data augmentation in video generation, it is limited to two strategies: frame-level shuffle and block-level shuffle. These methods, while effective, represent only an initial step in this direction. Future work could explore more advanced temporal augmentation techniques, such as motion-aware or context-sensitive strategies, to further enhance temporal coherence and diversity. We hope this study inspires broader research into temporal augmentations, paving the way for more robust and expressive video generation models.


\section{More Experimental Results}
\label{app_results}

\begin{figure*}[t]
    \centering
   \includegraphics[width=0.82\linewidth]{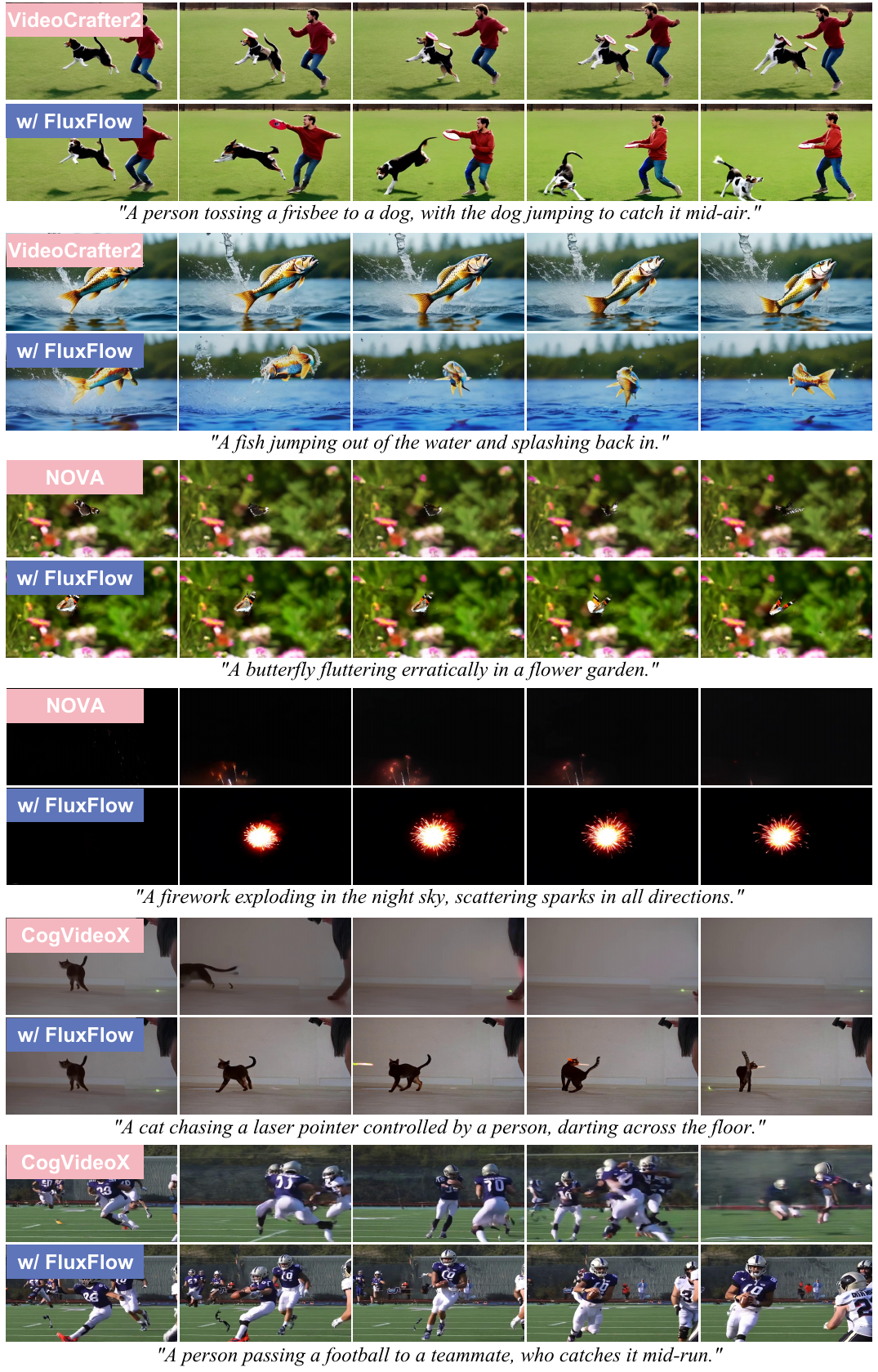}
   \vspace{-0.8em}
    \captionof{figure}{\label{fig:app_results}
    More comparison of \textsc{FluxFlow} on VideoCrafter2~\cite{chen2024videocrafter2}, NOVA~\cite{deng2024autoregressive}, and CogVideoX~\cite{yang2024cogvideox}.}
    \vspace{-0.8em}
\end{figure*}

We provide more comparison results here in Figure~\ref{fig:app_results}.

\end{document}